\documentclass[10pt,twocolumn,letterpaper]{article}

\usepackage{array,multirow}
\usepackage{cvpr}
\usepackage{times}
\usepackage{epsfig}
\usepackage{graphicx}
\usepackage{wrapfig}
\usepackage{amsmath}
\usepackage{amssymb}
\usepackage{xcolor}
\usepackage{caption}
\usepackage[utf8]{inputenc}

\usepackage{mathtools}
\usepackage{algorithm}
\usepackage[noend]{algpseudocode}

\usepackage[pagebackref=true,breaklinks=true,letterpaper=true,colorlinks,bookmarks=false]{hyperref}

\cvprfinalcopy 

\def\cvprPaperID{2172}

\DeclarePairedDelimiter\ceil{\lceil}{\rceil}

\newcommand{\x}{\mathbf{x}}
\newcommand{\y}{\mathbf{y}}
\newcommand{\m}{\mathbf{m}}

\newcommand{\z}{\mathbf{z}}
\newcommand{\R}{\mathbb{R}}

\begin{document}

\title{Conditional Probability Models for Deep Image Compression}

\author{
    \begin{tabular}{ccccc}
        Fabian Mentzer\thanks{The first two authors contributed equally.} &
        Eirikur Agustsson$^{*}$ &
        Michael Tschannen &
        Radu Timofte &
        Luc Van Gool\\
        {\scriptsize mentzerf@vision.ee.ethz.ch} &
        {\scriptsize aeirikur@vision.ee.ethz.ch} &
        {\scriptsize michaelt@nari.ee.ethz.ch} &
        {\scriptsize timofter@vision.ee.ethz.ch} &
        {\scriptsize vangool@vision.ee.ethz.ch} 
    \end{tabular}\\[3ex]
    ETH Z\"urich, Switzerland\vspace{-1ex}
}

\maketitle

\begin{abstract}
Deep Neural Networks trained as image auto-encoders have recently emerged as a promising direction for advancing the state-of-the-art in image compression. 
The key challenge in learning such networks is twofold: To deal with quantization, and to control the trade-off between reconstruction error (distortion) and entropy (rate) of the latent image representation. 
In this paper, we focus on the latter challenge and propose a new technique to navigate the rate-distortion trade-off for an image compression auto-encoder.
The main idea is to directly model the entropy of the latent representation by using a context model: A 3D-CNN which learns a conditional probability model of the latent distribution of the auto-encoder.
During training, the auto-encoder makes use of the context model to estimate the entropy of its representation, and the context model is concurrently updated to learn the dependencies between the symbols in the latent representation.
Our experiments show that this approach, when measured in MS-SSIM, yields a state-of-the-art image compression system based on a simple convolutional auto-encoder.
\end{abstract}

\section{Introduction }
\label{sec:introduction}

\begin{figure}[th!]
\centering
\includegraphics[width=0.9\linewidth]{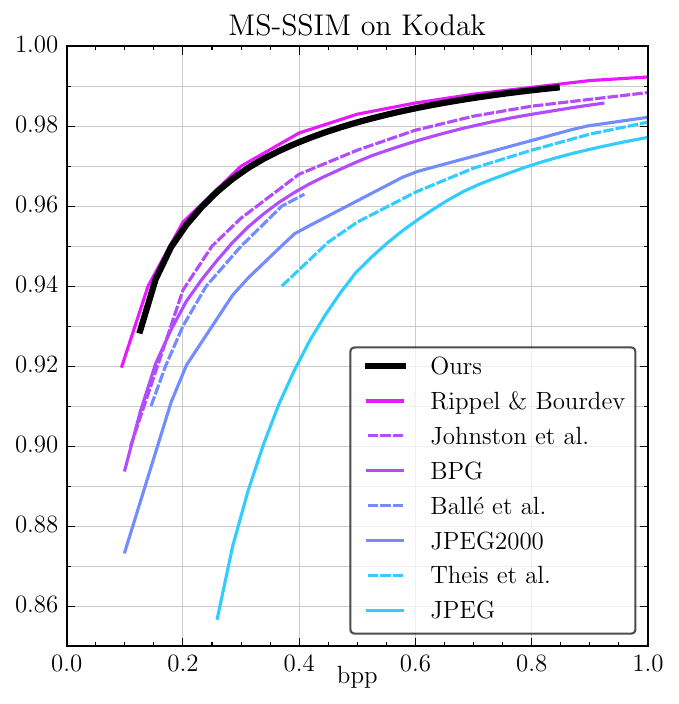}
\vspace{-0.4cm}
\caption{State-of-the-art performance achieved by our simple compression system composed of a standard convolutional auto-encoder and a 3D-CNN-based context model.}
\vspace{-0.4cm}
\label{fig:mean_kodak}
\end{figure}

Image compression refers to the task of representing images using as little storage (i.e., bits) as possible. While in lossless image compression the compression rate is limited by the requirement that the original image should be perfectly reconstructible, in \textit{lossy} image compression, a greater reduction in storage is enabled by allowing for some distortion in the reconstructed image. This results in a so-called rate-distortion trade-off, where a balance is found between the bitrate $R$ and the distortion $d$ by minimizing $d+\beta R$, where $\beta>0$ balances the two competing objectives.
Recently, deep neural networks (DNNs) trained as image auto-encoders for this task led to promising results, achieving better performance than many traditional techniques for image compression \cite{toderici2015variable, toderici2016full, theis2017lossy, balle2016end, agustsson2017soft, li2017learning}. Another advantage of DNN-based learned compression systems is their adaptability to specific target domains such as areal images or stereo images, enabling even higher compression rates on these domains. A key challenge in training such systems is to optimize the bitrate $R$ of the latent image representation in the auto-encoder. 
To encode the latent representation using a finite number of bits, it needs to be discretized into symbols (i.e., mapped to a stream of elements from some finite set of values).
Since discretization is non-differentiable, this presents  challenges for gradient-based optimization methods and many techniques have been proposed to address them.
After discretization, information theory tells us that the correct measure for bitrate $R$ is the entropy $H$ of the resulting symbols. 
Thus the challenge, and the focus of this paper, is how to model $H$ such that we can navigate the trade-off  $d+\beta H$ during optimization of the auto-encoder.

Our proposed method is based on leveraging context models, which were previously used as techniques to improve coding rates for already-trained models \cite{balle2016end, toderici2016full,li2017learning,rippel17a}, directly as an entropy term in the optimization. We concurrently train the auto-encoder and the context model with respect to each other, where the context model learns a convolutional probabilistic model of the image representation in the auto-encoder, while the auto-encoder uses it for entropy estimation to navigate the rate-distortion trade-off.
Furthermore, we generalize our formulation to spatially-aware networks, which use an \textit{importance map} to spatially attend the bitrate representation to the most important regions in the compressed representation.
The proposed techniques lead to a simple image compression system\footnote{\url{https://github.com/fab-jul/imgcomp-cvpr}},
which achieves state-of-the-art performance when measured with the popular multi-scale structural similarity index (MS-SSIM) distortion metric \cite{SSIM-MS}, while being straightforward to implement with standard deep-learning toolboxes.

\section{Related work}

Full-resolution image compression using DNNs has attracted considerable attention recently. DNN architectures commonly used for image compression are auto-encoders \cite{theis2017lossy, balle2016end, agustsson2017soft, li2017learning} and recurrent neural networks (RNNs) \cite{toderici2015variable, toderici2016full}. The networks are typically trained to minimize the mean-squared error (MSE) between original and decompressed image \cite{theis2017lossy, balle2016end, agustsson2017soft, li2017learning}, or using perceptual metrics such as MS-SSIM \cite{toderici2016full, rippel17a}. Other notable techniques involve progressive encoding/decoding strategies \cite{toderici2015variable, toderici2016full}, adversarial training \cite{rippel17a}, multi-scale image decompositions \cite{rippel17a}, and generalized divisive normalization (GDN) layers \cite{balle2016end,balle2016code}.

Context models and entropy estimation---the focus of the present paper---have a long history in the context of engineered compression methods, both lossless and lossy \cite{context1,context2,context3,context4,context5}.
Most of the recent DNN-based lossy image compression approaches have also employed such techniques in some form. \cite{balle2016end} uses a binary context model for adaptive binary arithmetic coding \cite{marpe2003context}. The works of \cite{toderici2016full,li2017learning,rippel17a} use learned context models for improved coding performance on their trained models when using adaptive arithmetic coding. \cite{theis2017lossy,agustsson2017soft} use non-adaptive arithmetic coding but estimate the entropy term with an independence assumption on the symbols.

Also related is the work of van den Oord \etal\cite{pixelrnn, pixelcnn}, who proposed PixelRNN and PixelCNN, powerful RNN- and CNN-based context models for modeling the distribution of natural images in a lossless setting, which can be used for (learned) lossless image compression as well as image generation.

\section{Proposed method}
\label{sec:proposed_method}
Given a set of training images $\mathcal{X}$, we wish to learn a compression system which consists of an encoder, a quantizer, and a decoder. 
The encoder $E:\R^d \to \R^m$ maps an image $\x$ to a latent representation $\z=E(\x)$. 
The quantizer $Q:\R \to \mathcal{C}$ discretizes the coordinates of $\z$ to $L=|\mathcal{C}|$ centers, obtaining $\hat{\z}$ with $\hat{z}_i:=Q(z_i)\in \mathcal{C}$, which can be losslessly encoded into a bitstream. 
The decoder $D$ then forms the reconstructed image $\hat{\x}=D(\hat{\z})$ from the quantized latent representation $\hat{\z}$, which is in turn (losslessy) decoded from the bitstream.
We want the encoded representation $\hat{\z}$ to be compact when measured in bits, while at the same time we want the distortion $d(\x,\hat{\x})$ to be small, where $d$ is some measure of reconstruction error, such as MSE or MS-SSIM.
This results in the so-called rate-distortion trade-off
\begin{equation}
d(\x,\hat{\x}) + \beta H(\hat{\z}), \label{eq:rd_tradeoff}
\end{equation}
where $H$ denotes the cost of encoding $\hat{\z}$ to bits, i.e., the entropy of $\hat{\z}$. Our system is realized by modeling $E$ and $D$ as convolutional neural networks (CNNs) (more specifically, as the encoder and decoder, respectively, of a convolutional auto-encoder) and minimizing \eqref{eq:rd_tradeoff} over the training set $\mathcal{X}$, where a large/small $\beta$ draws the system towards low/high average entropy $H$. 
In the next sections, we will discuss how we quantize $\z$ and estimate the entropy $H(\hat \z)$.
We note that as $E,D$ are CNNs, $\hat{\z}$ will be a 3D feature map, but for simplicity of exposition we will denote it as a vector with equally many elements. Thus, $\hat{z}_i$ refers to the $i$-th element of the feature map, in raster scan order (row by column by channel).

\subsection{Quantization}
We adopt the scalar variant of the quantization approach proposed in \cite{agustsson2017soft} to quantize $\z$, but simplify it using ideas from \cite{theis2017lossy}. Specifically, given centers $\mathcal{C}=\{c_1,\cdots,c_L\} \subset \R$, we use nearest neighbor assignments to compute
\begin{equation} \label{eq:hardquant}
\hat{z}_i = Q(z_i) := \text{arg min}_{j} \|z_i-c_j\|,
\end{equation}
but rely on (differentiable) soft quantization 
\begin{equation} \label{eq:softquant}
\tilde{z}_i = \sum_{j=1}^L \frac{\exp(-\sigma\|z_i-c_j\|)}{\sum_{l=1}^L \exp(-\sigma\|z_i-c_l\|)} c_j
\end{equation}
to compute gradients during the backward pass. This combines the benefit of \cite{agustsson2017soft} where the quantization is restricted to a finite set of learned centers $\mathcal{C}$ (instead of the fixed (non-learned) integer grid as in \cite{theis2017lossy}) and the simplicity of \cite{theis2017lossy}, where a differentiable approximation of quantization is only used in the backward pass, avoiding the need to choose an annealing strategy (i.e., a schedule for $\sigma$) as in \cite{agustsson2017soft} to drive the soft quantization \eqref{eq:softquant} to hard assignments \eqref{eq:hardquant} during training.
In TensorFlow, this is implemented as 
\begin{equation}
\bar{z}_i = \text{tf.stopgradient}(\hat{z}_i-\tilde{z}_i) + \tilde{z}_i. \label{eq:qbar}
\end{equation}
We note that for forward pass computations, $\bar{z}_i=\hat{z}_i$, and thus we will continue writing $\hat{z}_i$ for the latent representation.

\subsection{Entropy estimation}
\label{sec:entropyest}
To model the entropy $H(\hat{\z})$ we build on the approach of PixelRNN \cite{pixelrnn} and factorize the distribution $p(\hat{\z})$ as a product of conditional distributions
\begin{equation}
p(\hat{\z}) = \prod_{i=1}^m p(\hat{z}_i|\hat{z}_{i-1},\ldots,\hat{z}_1), \label{eq:condprob}
\end{equation}
where the 3D feature volume $\hat{\z}$ is indexed in raster scan order. 
We then use a neural network $P(\hat{\z})$, which we refer to as a \textit{context model}, to estimate each term $p(\hat{z}_i|\hat{z}_{i-1},\ldots,\hat{z}_1)$:
\begin{equation}
P_{i,l}(\hat{\z}) \approx p(\hat{z}_i=c_l|\hat{z}_{i-1},\ldots,\hat{z}_1),
\end{equation}
where $P_{i,l}$ specifies for every 3D location $i$ in $\hat\z$ the probabilites of each symbol in $\mathcal{C}$ with $l=1,\ldots,L$.
We refer to the resulting approximate distribution as $q(\hat{\z}):=\prod_{i=1}^m P_{i,I(\hat{z}_i)}(\hat{\z})$, where $I(\hat{z}_i)$ denotes the index of $\hat{z}_i$ in $\mathcal{C}$.

Since the conditional distributions $p(\hat{z}_i|\hat{z}_{i-1},\ldots,\hat{z}_1)$ only depend on previous values $\hat{z}_{i-1},\ldots,\hat{z}_1$, this imposes a \textit{causality} constraint on the network $P$: While $P$ may compute $P_{i,l}$ in parallel for $i=1,\ldots,m, l=1,\ldots,L$, it needs to make sure that each such term only depends on previous values $\hat{z}_{i-1},\ldots,\hat{z}_1$.

The authors of PixelCNN \cite{pixelrnn,pixelcnn} study the use of 2D-CNNs as causal conditional models over 2D images in a lossless setting, i.e., treating the RGB pixels as symbols. They show that the causality constraint can be efficiently enforced using masked filters in the convolution. Intuitively, the idea is as follows: If for each layer the causality condition is satisfied with respect to the spatial coordinates of the layer before, then by induction the causality condition will hold between the output layer and the input.
Satisfying the causality condition for each layer can be achieved with proper masking of its weight tensor, and thus the entire network can be made causal \textit{only through the masking of its weights}. Thus, the entire set of probabilities $P_{i,l}$ for all (2D) spatial locations $i$ and symbol values $l$ can be computed in parallel with a fully convolutional network, as opposed to modeling each term $p(\hat{z}_i|\hat{z}_{i-1},\cdots,\hat{z}_1)$ separately. 

In our case, $\hat{\z}$  is a 3D symbol volume, with as much as $K=64$ channels. We therefore generalize the approach of PixelCNN to 3D convolutions, using the same idea of masking the filters properly in every layer of the network.  This enables us to model $P$ efficiently, with a light-weight\footnote{We use a 4-layer network, compared to 15 layers in \cite{pixelrnn}.} 3D-CNN which slides over $\hat{z}$, while properly respecting the causality constraint. We refer to the supplementary material for more details.

As in \cite{pixelcnn}, we learn $P$ by training it for maximum likelihood, or equivalently (see \cite{shalizi06}) by training $P_{i,:}$  to classify the index $I(\hat{z}_i)$ of $\hat{z}_i$ in $\mathcal{C}$ with a cross entropy loss:
\begin{equation}
CE := \mathbb{E}_{\hat{z}\sim p(\hat{\z})}[\sum_{i=1}^m -\log P_{i,I(\hat{z}_i)}]. \label{eq:crossent}
\end{equation}
Using the well-known property of cross entropy as the coding cost when using the wrong distribution $q(\hat{\z})$ instead of the true distribution $p(\hat{\z})$, we can also view the $CE$ loss as an estimate of $H(\hat{\z})$ since we learn $P$ such that $P=q\approx p$.
That is, we can compute 
\begin{align}
H(\hat{\z}) &= \mathbb{E}_{\hat{z}\sim p(\hat{\z})}[-\log(p(\hat{\z}))]\label{eq:condent}\\
&= \mathbb{E}_{\hat{z}\sim p(\hat{\z})}[\sum_{i=1}^m -\log p(\hat{z}_i|\hat{z}_{i-1},\cdots,\hat{z}_1)]\label{eq:condent_sum}\\
&\approx \mathbb{E}_{\hat{z}\sim p(\hat{\z})}[\sum_{i=1}^m -\log q(\hat{z}_i|\hat{z}_{i-1},\cdots,\hat{z}_1)]\\
&= \mathbb{E}_{\hat{z}\sim p(\hat{\z})}[\sum_{i=1}^m -\log P_{i,I(\hat{z}_i)}]\\
&= CE \label{eq:crossent2}
\end{align}
Therefore, when training the auto-encoder we can indirectly minimize $H(\hat{\z})$ through the cross entropy $CE$.
We refer to argument in the expectation of \eqref{eq:crossent}, 
\begin{equation}
C(\hat{\z}) := \sum_{i=1}^m -\log P_{i,I(\hat{z}_i)}\label{eq:coding_cost},
\end{equation}
 as the \emph{coding cost} of the latent image representation, since this reflects the coding cost incurred when using $P$ as a context model with an adaptive arithmetic encoder~\cite{marpe2003context}.
 From the application perspective, minimizing the coding cost is actually more important than the (unknown) true entropy, since it reflects the bitrate obtained in practice.
 
To backpropagate through $P(\hat{\z})$ we use the same approach as for the encoder (see \eqref{eq:qbar}). Thus, like the decoder $D$, $P$ only sees the (discrete) $\hat{\z}$ in the forward pass, whereas the gradient of the soft quantization $\tilde{\z}$ is used for the backward pass.

\subsection{Concurrent optimization}
Given an auto-encoder $(E,D)$, we can train $P$ to model the dependencies of the entries of $\hat{\z}$ as described in the previous section by minimizing \eqref{eq:crossent}. 
On the other hand, using the model $P$, we can obtain an  estimate of $H(\hat{\z})$ as in \eqref{eq:crossent2} and use this estimate to adjust $(E,D)$ such that $d(\x,D(Q(E(\x))))+\beta H(\hat{\z})$ is reduced, thereby navigating the rate distortion trade-off. 
Therefore, it is natural to concurrently learn $P$ (with respect to its own loss), and $(E,D)$  (with respect to the rate distortion trade-off) during training, such that all models which the losses depend on are continuously updated.

\subsection{Importance map for spatial bit-allocation}
\label{sec:importancemap}
Recall that since $E$ and $D$ are CNNs, $\hat{\z}$ is a 3D feature-map.
For example, if $E$ has three stride-2 convolution layers and the bottleneck has $K$ channels, the dimensions of $\hat{\z}$ will be $\frac{W}{8}\times \frac{H}{8}\times K$. A consequence of this formulation is that we are using equally many symbols in $\hat{\z}$ for each spatial location of the input image $\x$. It is known, however, that in practice there is great variability in the information content across spatial locations (e.g., the uniform area of blue sky vs. the fine-grained structure of the leaves of a tree).

This can in principle be accounted for automatically in the trade-off between the entropy and the distortion, where the network would learn to output more predictable (i.e., low entropy) symbols for the low information regions, while making room for the use of high entropy symbols for the more complex regions.
More precisely, the formulation in \eqref{eq:crossent} already allows for variable bit allocation for different spatial regions through the context model $P$.

However, this arguably requires a quite sophisticated (and hence computationally expensive) context model, and we find it beneficial to follow Li \etal \cite{li2017learning} instead by using an \textit{importance map} to help the CNN attend to different regions of the image with different amounts of bits.
While \cite{li2017learning} uses a separate network for this purpose, we consider a simplified setting. We take the last layer of the encoder $E$, and add a second single-channel output $\y\in \R^{\frac{W}{8}\times \frac{H}{8}\times 1}$. 
We expand this single channel $\y$ into a mask $\m\in\R^{\frac{W}{8}\times \frac{H}{8}\times K}$ of the same dimensionality as $\z$ as follows:
\begin{equation}
m_{i,j,k} = \begin{cases}
1  & \text{ if } k < y_{i,j}\\
(y_{i,j}-k)  & \text{ if }  k \leq y_{i,j} \leq k+1 \\
0  & \text{ if } k+1 > y_{i,j} \\
\end{cases},\label{eq:expand_map}
\end{equation}
 where $y_{i,j}$ denotes the value of $\y$ at spatial location $(i,j)$.
The transition value for $k \leq y_{i,j} \leq k+1$ is such that the mask smoothly transitions from 0 to 1 for non-integer values of $\y$.

We then mask $\z$ by pointwise multiplication with the binarization of $\m$, i.e., $\z \leftarrow \z \odot \lceil \m \rceil$.
Since the ceiling operator $\lceil \cdot \rceil$ is not differentiable, as done by \cite{theis2017lossy,li2017learning}, we use identity for the backward pass.

With this modification, we have simply changed the architecture of $E$ slightly such that it can easily ``zero out'' portions of columns $\z_{i,j,:}$ of $\z$ (the rest of the network stays the same, so that \eqref{eq:hardquant} still holds for example).
As suggested by \cite{li2017learning}, the so-obtained structure in $\z$ presents an alternative coding strategy: Instead of losslessly encoding the entire symbol volume $\hat{\z}$, we could first (separately) encode the mask $\lceil \m \rceil$, and then  for each column $\hat{z}_{i,j,:}$ only encode the first $\lceil m_{i,j}\rceil+1$ symbols, since the remaining ones are the constant $Q(0)$, which we refer to as the \textit{zero symbol}.

Work \cite{li2017learning} uses binary symbols (i.e., $\mathcal{C}=\{0,1\}$) and assumes independence between the symbols and a uniform prior during training, i.e., costing each 1 bit to encode. The importance map is thus their principal tool for controlling the bitrate, since they thereby avoid encoding all the bits in the representation. In contrast, we stick to the formulation in \eqref{eq:condprob} where the dependencies between the symbols are modeled during training. We then use the importance map as an architectural constraint and use their suggested coding strategy to obtain an alternative estimate for the entropy $H(\hat{\z})$, as follows.

We observe that we can recover $\lceil \m \rceil$ from $\hat{\z}$ by counting the number of consecutive zero symbols at the end of each column $\hat{\z}_{i,j,:}$.\footnote{If $\z$ contained zeros before it was masked, we might overestimate the number of $0$ entries in $\lceil \m \rceil$. However, we can redefine those entries of $\m$ as $0$ and this will give the same result after masking. } $\lceil \m \rceil$ is therefore a function of the masked $\hat{\z}$, i.e., $\lceil \m \rceil=g(\hat{\z})$ for $g$ recovering $\lceil \m \rceil$ as described, which means that we have for the conditional entropy $H(\lceil \m \rceil|\hat{\z})=0$. Now, we have
\begin{align}
H(\hat{\z}) &= H(\lceil \m \rceil|\hat{\z}) +  H(\hat{\z})\\
&=H(\hat{\z},\lceil \m \rceil) \\
&= H(\hat{\z}|\lceil \m \rceil) + H(\lceil \m \rceil).
\end{align}
If we treat the entropy of the mask, $H(\lceil \m \rceil)$, as constant during optimization of the auto-encoder, we can then indirectly minimize $H(\hat{\z})$ through $H(\hat{\z}|\m)$.

To estimate $H(\hat{\z}|\m)$, we use the same factorization of $p$ as in \eqref{eq:condprob}, but since the mask $\lceil \m \rceil$ is known we have $p(\hat{z}_i=c_0)=1$ deterministic for the 3D locations $i$ in $\hat{\z}$ where the mask is zero.
The $\log$s of the corresponding terms in \eqref{eq:condent_sum} then evaluate to $0$. The remaining terms, we can model with the same context model $P_{i,l}(\hat{\z})$, which results in
\begin{equation}
H(\hat{\z}|\lceil \m \rceil) \approx E_{\hat{z}\sim p(\hat{\z})}[\sum_{i=1}^m -\lceil m_i \rceil \log P_{i,I(\hat{z}_i)}],\label{eq:croessent_weight}
\end{equation}
where $m_i$ denotes the $i$-th element of $\m$ (in the same raster scan order as $\hat{z}$).

Similar to the coding cost \eqref{eq:coding_cost}, we refer to the argument in the expectation in \eqref{eq:croessent_weight}, 
\begin{equation}
MC(\hat{\z}):=\sum_{i=1}^m -\lceil m_i \rceil \log P_{i,I(\hat{z}_i)}\label{eq:masked_coding_cost}
\end{equation}
as the \emph{masked coding cost} of $\hat{\z}$.

While the entropy estimate \eqref{eq:croessent_weight} is almost estimating the same quantity as \eqref{eq:crossent} (only differing by $H(\lceil \m \rceil)$), it has the benefit of being weighted by $m_i$. Therefore, the encoder $E$ has an obvious path to control the entropy of $\hat{\z}$, by simply increasing/decreasing the value of $\y$ for some spatial location of $\x$ and thus obtaining fewer/more zero entries in $\m$.

When the context model $P(\hat{\z})$ is trained, however, we still train it with respect to the formulation in \eqref{eq:condent}, so it does not have direct access to the mask $\m$ and needs to learn the dependencies on the entire masked symbol volume $\hat{\z}$. This means that when encoding an image, we can stick to standard adaptive arithmetic coding over the entire bottleneck, without needing to resort to a two-step coding process as in \cite{li2017learning}, where the mask is first encoded and then the remaining symbols. We emphasize that this approach hinges critically on the context model $P$ and the encoder $E$ being trained concurrently as this allows the encoder to learn a meaningful (in terms of coding cost) mask with respect to $P$ (see the next section).

In our experiments we observe that during training, the two entropy losses \eqref{eq:crossent} and \eqref{eq:croessent_weight} converge to almost the same value, with the latter being around $\approx 3.5\%$ smaller due to $H(\lceil \m \rceil)$ being ignored. 

While the importance map is not crucial for optimal rate-distortion performance, if the channel depth $K$ is adjusted carefully, we found that we could more easily control the entropy of $\hat{\z}$ through $\beta$ when using a fixed $K$, since the network can easily learn to ignore some of the channels via the importance map.  
Furthermore, in the supplementary material we show that by using multiple importance maps for a single network, one can obtain a single model that supports multiple compression rates.

\subsection{Putting the pieces together}
We made an effort to carefully describe our formulation and its motivation in detail. While the description is lengthy, when putting the resulting pieces together we get a quite straightforward pipeline for learned image compression, as follows.

Given the set of training images $\mathcal{X}$, we initialize (fully convolutional) CNNs $E$, $D$, and $P$, as well as the centers $\mathcal{C}$ of the quantizer $Q$. 
Then, we train over minibatches $\mathcal{X}_B=\{\x^{(1)},\cdots,\x^{(B)}\}$ of crops from $\mathcal{X}$.
At each iteration, we take one gradient step for the auto-encoder $(E,D)$ and the quantizer $Q$, with respect to the rate-distortion trade-off
\begin{equation}
\mathcal L_{E,D,Q} = \frac{1}{B}\sum_{j=1}^B d(\x^{(j)},\hat{\x}^{(j)})+\beta MC(\hat{\z}^{(j)}),\label{eq:rd_loss}
\end{equation}
which is obtained by combining \eqref{eq:rd_tradeoff} with the estimate \eqref{eq:croessent_weight} \& \eqref{eq:masked_coding_cost} and taking the batch sample average.
Furthermore, we take a gradient step for the context model $P$ with respect to its objective (see \eqref{eq:crossent} \&  \eqref{eq:coding_cost})
\begin{equation}
\mathcal L_P := \frac{1}{B} \sum_{j=1}^B d(\x^{(j)},\hat{\x}^{(j)})+\beta C(\hat{\z}^{(j)}).
\end{equation}

To compute these two batch losses, we need to perform the following computation for each  $\x\in\mathcal{X}_B$:
\begin{enumerate}
\setlength\itemsep{0.1em}
\item Obtain compressed (latent) representation $\z$ and importance map $\y$ from the encoder: $(\z,\y) = E(\x)$
\item Expand importance map $\y$ to mask $\m$ via \eqref{eq:expand_map}
\item Mask $\z$, i.e., $\z\leftarrow \z \odot \lceil \m \rceil$
\item Quantize $\hat{\z}=Q(\z)$
\item Compute the context $P(\hat{\z})$
\item Decode $\hat{\x}=D(\hat{\z})$,
\end{enumerate}
which can be computed in parallel over the minibatch on a GPU since all the models are fully convolutional.

\subsection{Relationship to previous methods}
We are not the first to use context models for adaptive arithmetic coding to improve the performance in learned deep image compression. Work \cite{toderici2016full} uses a PixelRNN-like architecture \cite{pixelrnn} to train a recurrent network as a context model for an RNN-based compression auto-encoder.
Li \etal \cite{li2017learning} extract cuboid patches around each symbol in a binary feature map, and feed them to a convolutional context model.   
Both these methods, however, only learn the context model \textit{after} training their system, as a post-processing step to boost coding performance. 

In contrast, our method directly incorporates the context model as the entropy term for the rate-distortion term \eqref{eq:rd_tradeoff} of the auto-encoder, and trains the two concurrently. This is done at little overhead during training, since we adopt a 3D-CNN for the context model, using PixelCNN-inspired \cite{pixelcnn} masking of the weights of each layer to ensure causality in the context model. 
Adopting the same approach to the context models deployed by \cite{toderici2016full} or \cite{li2017learning} would be non-trivial since they are not designed for fast feed-forward computation.  In particular, while the context model of \cite{li2017learning} is also convolutional, its causality is enforced through masking the \textit{inputs} to the network, as opposed to our masking of the weights of the networks. This means their context model needs to be run separately with a proper input cuboid for each symbol in the volume (i.e., not fully convolutionally).


\section{Experiments}
\label{sec:experiments}

\begin{figure}[htb!]
\centering
\includegraphics[width=1.02\linewidth]{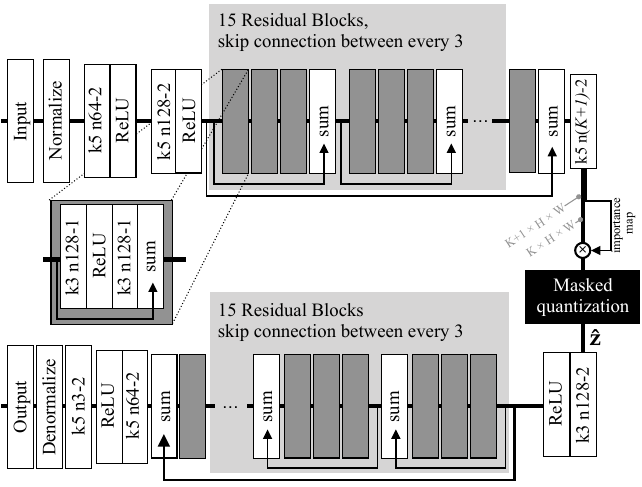}
\vspace{-0.3cm}
\caption{The architecture of our auto-encoder. Dark gray blocks represent residual units. The upper part represents the encoder $E$, the lower part the decoder $D$. For the encoder, ``$\text{k}5\text{ n}64\text{-}2$'' represents a convolution layer with kernel size 5, 64 output channels and a stride of 2. For the decoder it represents the equivalent deconvolution layer. All convolution layers are normalized using batch norm~\cite{ioffe2015batch}, and use SAME padding. \textit{Masked quantization} is the quantization described in Section~\ref{sec:importancemap}. \textit{Normalize} normalizes the input to $[0, 1]$ using a mean and variance obtained from a subset of the training set. \textit{Denormalize} is the inverse operation.}
\label{fig:archAE}
\vspace{-0.22cm}
\end{figure}

\begin{figure}[htb!]
\centering
\includegraphics[width=0.6\linewidth]{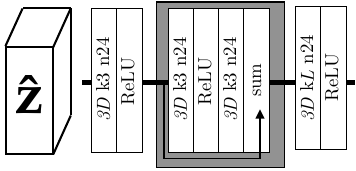}
\vspace{-0.15cm}
\caption{The architecture of our context model. ``$3D\text{ k}3\text{ n}24$'' refers to a 3D \textit{masked} convolution with filter size 3 and 24 output channels. The last layer outputs $L$ values for each voxel in $\hat \z$.}
\label{fig:archPC}
\vspace{-0.3cm}
\end{figure}

\paragraph{Architecture} Our auto-encoder has a similar architecture as~\cite{theis2017lossy} but with more layers, and is described in Fig.~\ref{fig:archAE}. We adapt the number of channels $K$ in the latent representation for different models.
For the context model $P$, we use a simple $4$-layer 3D-CNN as described in Fig.~\ref{fig:archPC}. 

\paragraph{Distortion measure} Following \cite{google_newpaper, rippel17a}, we use the multi-scale structural similarity index (MS-SSIM)~\cite{SSIM-MS} as measure of distortion $d(\x,\hat{\x})=100\cdot(1-\text{MS-SSIM}(\x,\hat{\x}))$ for our models. MS-SSIM reportedly correlates better with human perception of distortion than mean squared error (MSE). We train and test all our models using MS-SSIM.

\paragraph{Training} We use the Adam optimizer~\cite{kingmaB14} with a minibatch size of 30 to train seven models. Each model is trained to maximize MS-SSIM directly. As a baseline, we used a learning rate (LR) of $4 \cdot 10^{-3}$ for each model, but found it beneficial to vary it slightly for different models. We set $\sigma=1$ in the smooth approximation \eqref{eq:softquant} used for gradient backpropagation through $Q$. 
To make the model more predictably land at a certain bitrate $t$ when optimizing \eqref{eq:rd_tradeoff}, we found it helpful to clip the rate term (i.e., replace the entropy term $\beta H$ with $\max(t,\beta H)$), such that the entropy term is ``switched off'' when it is below $t$. We found this did not hurt performance.
We decay the learning rate by a factor 10 every two epochs. To obtain models for different bitrates, we adapt the target bitrate $t$ and the number of channels $K$, while using a moderately large $\beta=10$. We use a small regularization on the weights and note that we achieve very stable training. We trained our models for 6 epochs, which took around 24h per model on a single GPU.
For $P$, we use a LR of $10^{-4}$ and the same decay schedule.

\vspace{-0.2cm}
\paragraph{Datasets} We train on the the \textbf{ImageNet} dataset from the Large Scale Visual Recognition Challenge 2012 (ILSVRC2012)~\cite{ImageNETpaper}. As a preprocessing step, we take random $160 \times 160$ crops, and randomly flip them. We set aside 100 images from ImageNet as a testing set, \mbox{\textbf{ImageNetTest}}. Furthermore, we test our method on the widely used \textbf{Kodak}~\cite{kodakurl} dataset. To asses performance on high-quality full-resolution images, we also test on the datasets \textbf{B100}~\cite{Timofte2014} and \textbf{Urban100}~\cite{Huang-CVPR-2015}, commonly used in super-resolution.
\vspace{-0.2cm}
\paragraph{Other codecs} We compare to JPEG, using libjpeg\footnote{\url{http://libjpeg.sourceforge.net/}}, and JPEG2000, using the Kakadu implementation\footnote{\url{http://kakadusoftware.com/}}. We also compare to the lesser known BPG\footnote{\url{https://bellard.org/bpg/}}, which is based on HEVC, the state-of-the-art in video compression, and which outperforms JPEG and JPEG2000. We use BPG in the non-default 4:4:4 chroma format, following~\cite{rippel17a}.

\vspace{-0.2cm}
\paragraph{Comparison} Like~\cite{rippel17a}, we proceed as follows to compare to other methods. For each dataset, we compress each image using all our models. This yields a set of (bpp, MS-SSIM) points for each image, which we interpolate to get a curve for each image. We fix a grid of bpp values, and average the curves for each image at each bpp grid value (ignoring those images whose bpp range does not include the grid value, i.e., we do not extrapolate). We do this for our method, BPG, JPEG, and JPEG2000. 
Due to code being unavailable for the related works in general, we digitize the Kodak curve from Rippel \& Bourdev \cite{rippel17a}, who have carefully collected the curves from the respective works.
With this, we also show the results of Rippel \& Bourdev~\cite{rippel17a}, Johnston \etal~\cite{google_newpaper}, Ballé \etal~\cite{balle2016end}, and Theis \etal~\cite{theis2017lossy}.  To validate that our estimated MS-SSIM is correctly implemented, we independently generated the BPG curves for Kodak and verified that they matched the one from \cite{rippel17a}.

\begin{figure*}[thb!]
\vspace{-0.5cm}
\captionsetup{width=0.735\linewidth}
\centering
\setlength{\tabcolsep}{1pt}
\resizebox{0.8\linewidth}{!}
{
\begin{tabular}{ccc}
    \includegraphics[width=0.32\textwidth]{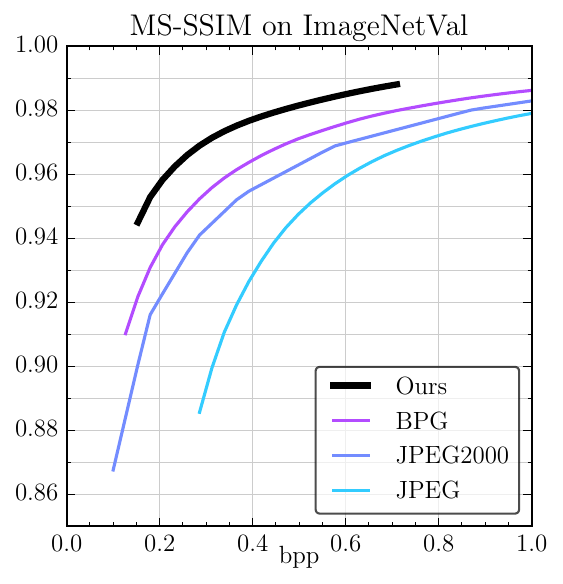}&
    \includegraphics[width=0.32\textwidth]{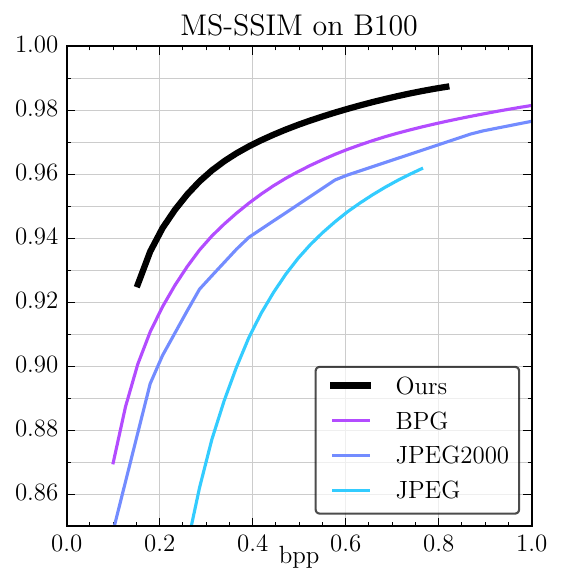}&
    \includegraphics[width=0.32\textwidth]{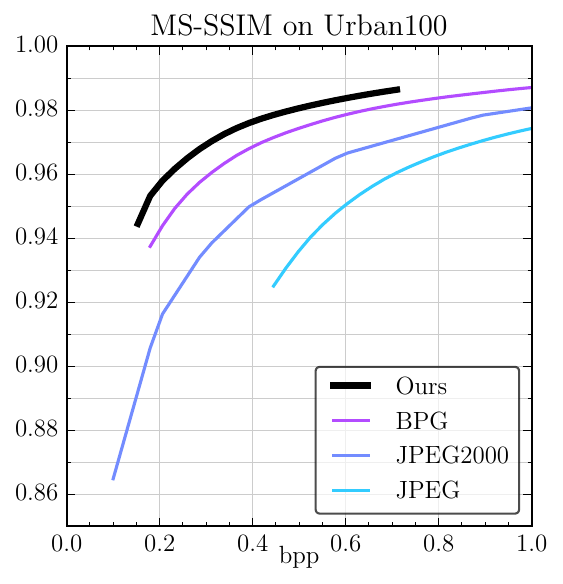}\\
\end{tabular}
}
\vspace{-0.5cm}
\caption{Performance of our approach on ImageNetTest, B100, Urban100, where we outperform BPG, JPEG and JPEG2000 in MS-SSIM.}
\label{fig:comp_imgnet_b100_u100}
\end{figure*}
\vspace{-0.2cm}

\paragraph{Results} Fig.~\ref{fig:mean_kodak} shows a comparison of the aforementioned methods for Kodak. Our method outperforms BPG, JPEG, and JPEG2000, as well as the neural network based approaches of Johnston \etal \cite{google_newpaper}, Ballé \etal \cite{balle2016end}, and Theis \etal \cite{theis2017lossy}. Furthermore, we achieve performance comparable to that of Rippel \& Bourdev~\cite{rippel17a}. This holds for all bpps we tested, from 0.3 bpp to 0.9 bpp. We note that while Rippel \& Bourdev and Johnston \etal also train to maximize (MS-)SSIM, the other methods minimize MSE.

\begin{figure*}[thb!]
\centering
\noindent\makebox[\textwidth]{
\resizebox{0.8\textwidth}{!}
{
\setlength{\tabcolsep}{1pt}
\begin{tabular}{lr}
\textbf{Ours} 0.124bpp&
0.147 bpp \textbf{BPG} \\
\includegraphics[width=0.5\textwidth]{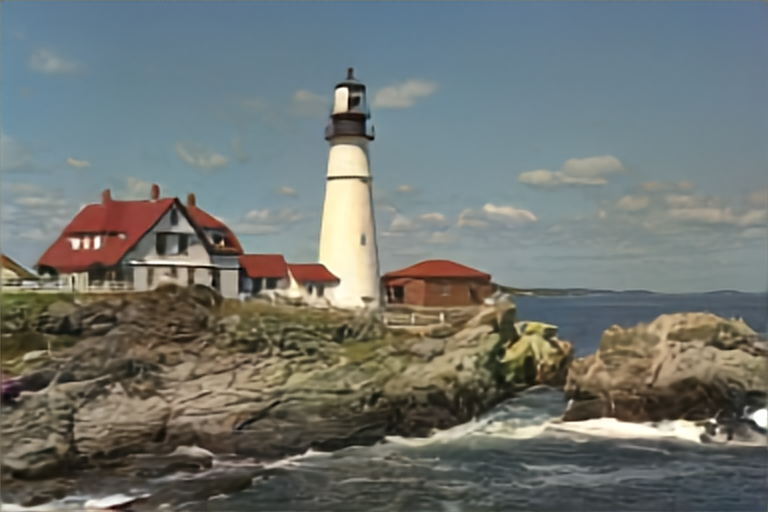}&
\includegraphics[width=0.5\textwidth]{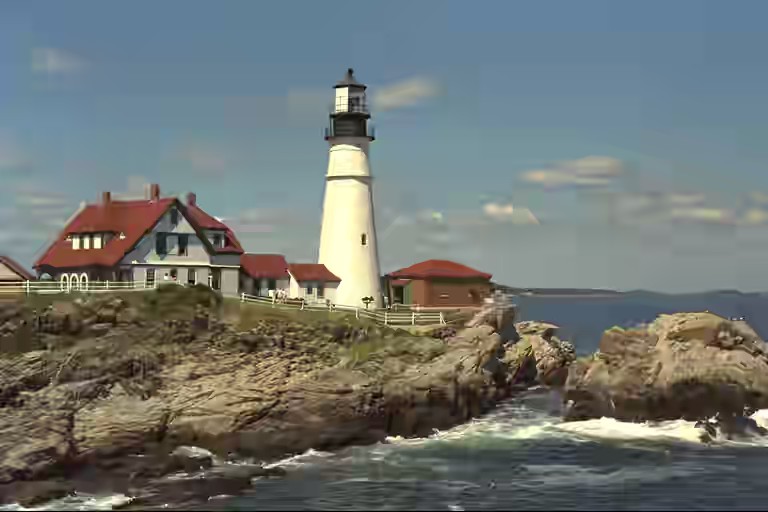}\\
\includegraphics[width=0.5\textwidth]{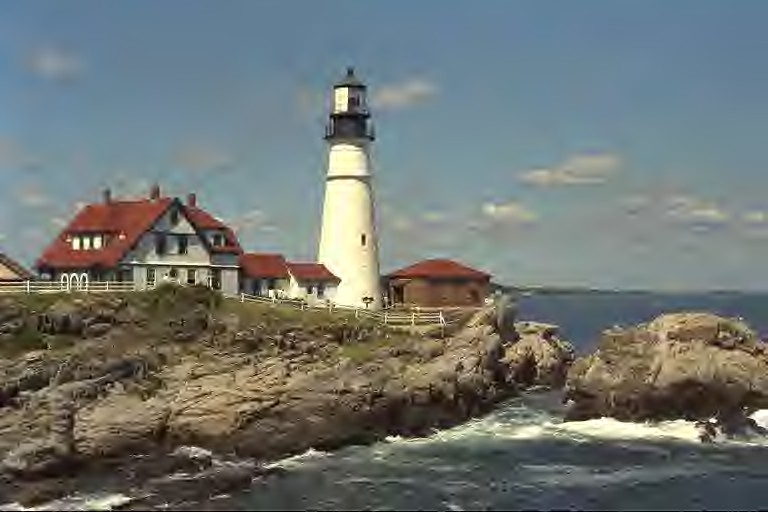}&
\includegraphics[width=0.5\textwidth]{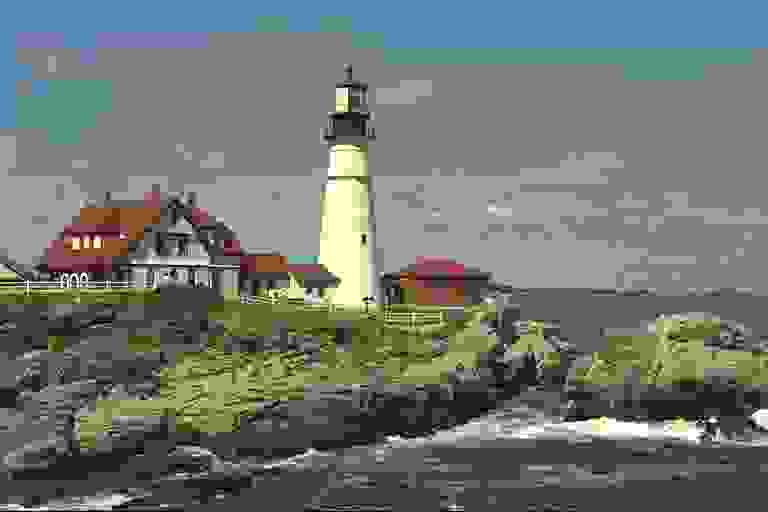}\\
\textbf{JPEG2000} 0.134bpp&
0.150bpp \textbf{JPEG}  \\
\end{tabular}
}
}
\vspace{-0.1cm}
\caption{Example image (\textit{kodim21}) from the Kodak testing set, compressed with different methods.}
\label{fig:kodim21}
\vspace{-0.3cm}
\end{figure*}

In each of the other testing sets, we also outperform BPG, JPEG, and JPEG2000 over the reported bitrates, as shown in Fig.~\ref{fig:comp_imgnet_b100_u100}. 

In Fig.~\ref{fig:kodim21}, we compare our approach to BPG, JPEG, and JPEG2000 visually, using very strong compression on \textit{kodim21} from Kodak. It can be seen that the output of our network is pleasant to look at. Soft structures like the clouds are very well preserved. BPG appears to handle high frequencies better (see, e.g., the fence) but loses structure in the clouds and in the sea. Like JPEG2000, it produces block artifacts. JPEG breaks down at this rate. We refer to the supplementary material for further visual examples.

\vspace{-0.3cm}
\paragraph{Ablation study: Context model} In order to show the effectiveness of the context model, we performed the following ablation study. We trained the auto-encoder without entropy loss, i.e., $\beta = 0$ in \eqref{eq:rd_loss}, using $L=6$ centers and $K=16$ channels. On Kodak, this model yields an average MS-SSIM of 0.982, at an average rate of 0.646 bpp (calculated assuming that we need $\log_2(L) = 2.59$ bits per symbol). We then trained three different context models for this auto-encoder, while keeping the auto-encoder fixed: A \textit{zeroth order} context model which uses a histogram to estimate the probability of each of the $L$ symbols; a \textit{first order} (one-step prediction) context model, which uses a conditional histogram to estimate the probability of each of the $L$ symbols given the previous symbol (scanning $\hat \z$ in raster order); and $P$, i.e., our proposed context model. The resulting average rates are shown in Table~\ref{tab:ablationcm}. Our context model reduces the rate by 10 \%, even though the auto-encoder was optimized using a uniform prior (see supplementary material for a detailed comparison of Table~\ref{tab:ablationcm} and Fig.~\ref{fig:mean_kodak}).

\begin{table}[h]
\centering
\vspace{-0.2cm}
\setlength{\tabcolsep}{3pt}
\begin{tabular}{l l l}
	Model & rate \\
	\hline
    Baseline (Uniform)    & 0.646 bpp \\
    Zeroth order            & 0.642 bpp \\
    First order           & 0.627 bpp \\
    \hline
    Our context model $P$ & 0.579 bpp 
\end{tabular}
\vspace{-0.1cm}
\caption{\label{tab:ablationcm} Rates for different context models, for the same architecture $(E,D)$.}
\vspace{-0.5cm}
\end{table}
\vspace{-0.3cm}
\paragraph{Importance map} As described in detail in Section~\ref{sec:importancemap}, we use an importance map to dynamically alter the number of channels used at different spatial locations to encode an image. To visualize how this helps, we trained two auto-encoders $M$ and $M'$, where $M$ uses an importance map and at most $K=32$ channels to compress an image, and $M'$ compresses without importance map and with $K=16$ channels (this yields a rate for $M'$ similar to that of $M$). In Fig.~\ref{fig:vis_im},
we show an image from ImageNetTest along with the same image compressed to 0.463 bpp by $M$ and compressed to 0.504 bpp by $M'$. Furthermore, Fig.~\ref{fig:vis_im} shows the importance map produced by $M$, as well as ordered visualizations of all channels of the latent representation for both $M$ and $M'$. Note how for $M$, channels with larger index are sparser, showing how the model can spatially adapt the number of channels. $M'$ uses all channels similarly.

\begin{figure}[h]
\vspace{-0.25cm}

\centering

\resizebox{\linewidth}{!}
{
\setlength{\tabcolsep}{1pt}
\begin{tabular}{cc}
    Input & Importance map of $M$ \\
    \includegraphics[width=0.5\linewidth]{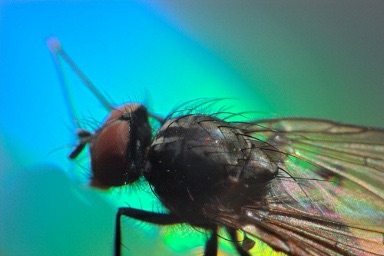}&
    \includegraphics[width=0.5\linewidth]{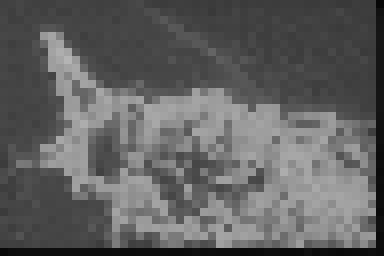}\\
     Output of $M$ & Latent representation of $M$ \\
    \includegraphics[width=0.5\linewidth]{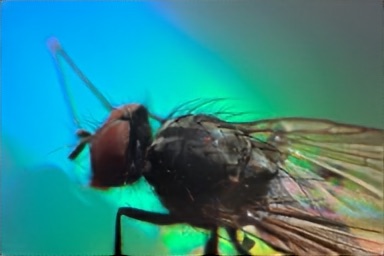}&
    \includegraphics[width=0.5\linewidth]{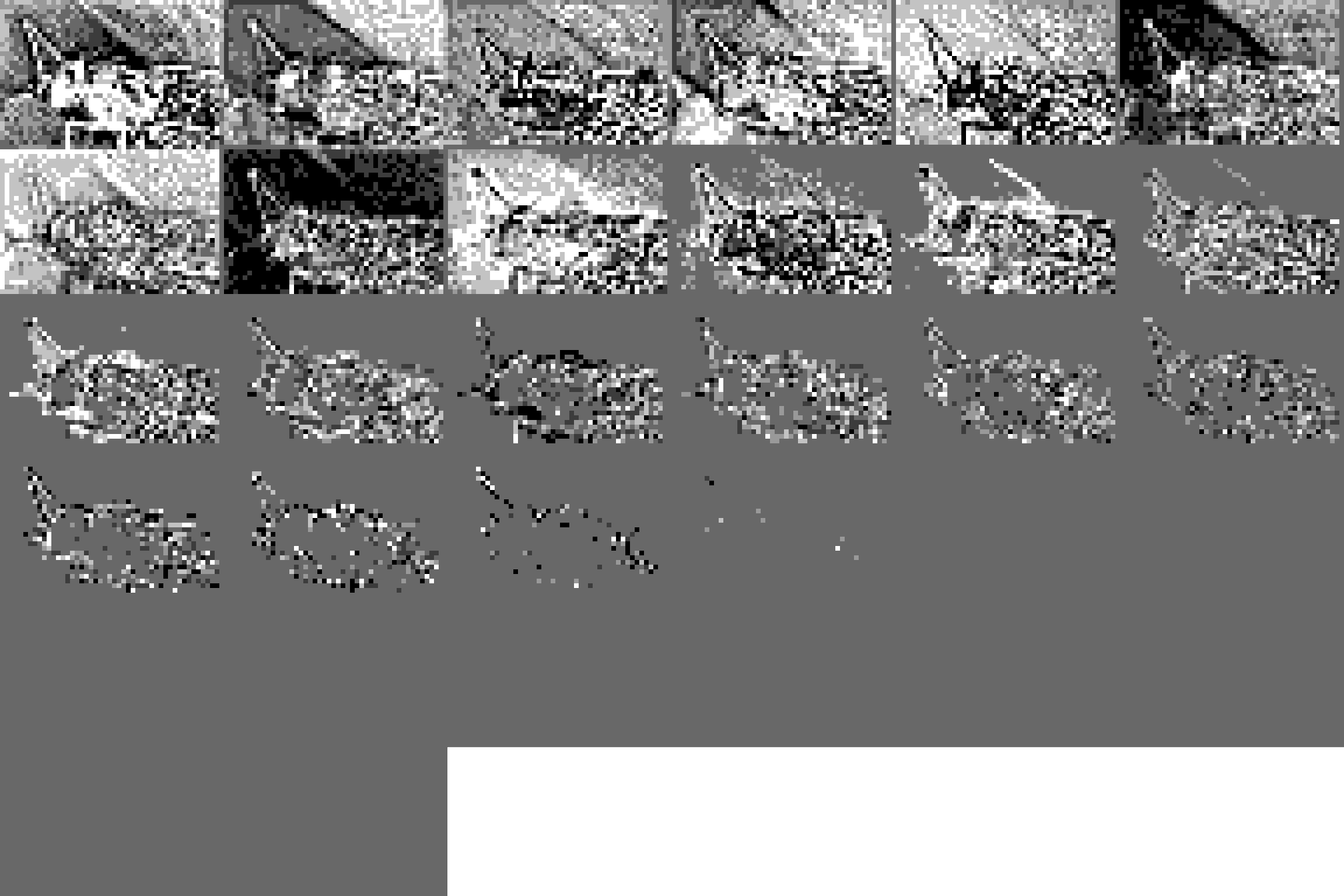}\\
     Output of $M'$ & Latent representation of $M'$ \\
    \includegraphics[width=0.5\linewidth]{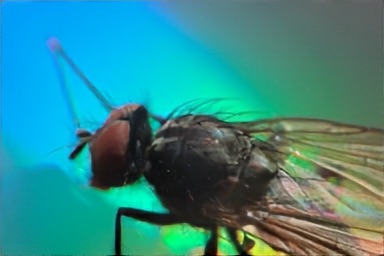}&
    \includegraphics[width=0.5\linewidth]{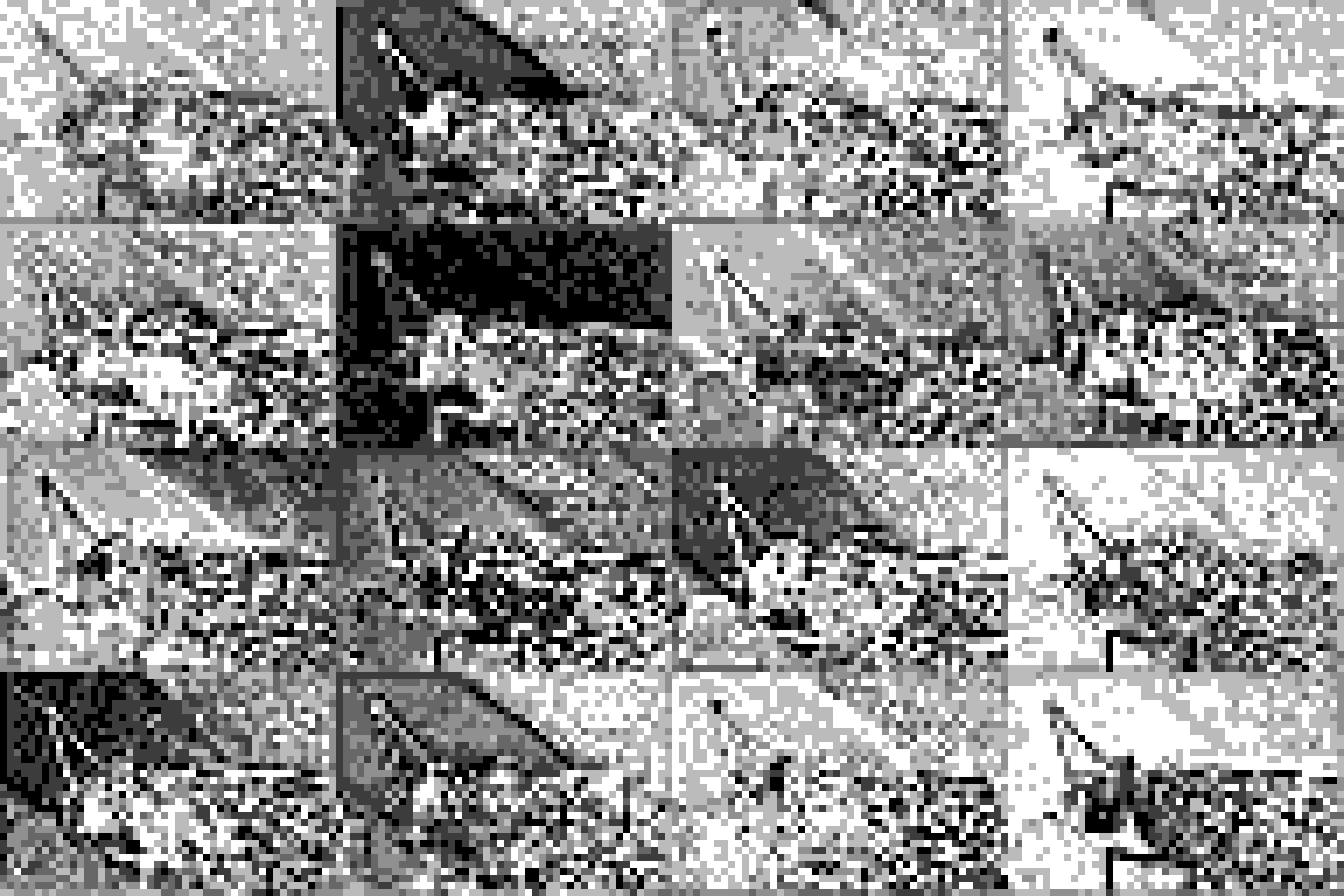}\\
\end{tabular}
}
\vspace{-0.3cm}
\caption{Visualization of the latent representation of the auto-encoder for a high-bpp operating point, with ($M$) and without ($M'$) incorporating an importance map.}
\label{fig:vis_im}
\vspace{-0.5cm}
\end{figure}

\vspace{-0.1cm}
\section{Discussion}

Our experiments showed that combining a convolutional auto-encoder with a lightweight 3D-CNN as context model and training the two networks concurrently leads to a highly effective image compression system. Not only were we able to clearly outperform state-of-the-art engineered compression methods including BPG and JPEG2000 in terms of MS-SSIM, but we also obtained performance competitive with the current state-of-the-art learned compression method from \cite{rippel17a}. 
In particular, our method outperforms BPG and JPEG2000 in MS-SSIM across four different testing sets (ImageNetTest, Kodak, B100, Urban100), and does so significantly, i.e., the proposed method generalizes well. 
We emphasize that our method relies on elementary techniques both in terms of the architecture (standard convolutional auto-encoder with importance map, convolutional context model) and training procedure (minimize the rate-distortion trade-off and the negative log-likelihood for the context model), while \cite{rippel17a} uses highly specialized techniques such as a pyramidal decomposition architecture,
adaptive codelength regularization, and multiscale adversarial training. 

The ablation study for the context model showed that our 3D-CNN-based context model is significantly more powerful than the first order (histogram) and second order (one-step prediction) baseline context models. Further, our experiments suggest that the importance map learns to condensate the image information in a reduced number of channels of the latent representation without relying on explicit supervision. 
Notably, the importance map is learned as a part of the image compression auto-encoder concurrently with the auto-encoder and the context model, without introducing any optimization difficulties. In contrast, in \cite{li2017learning} the importance map is computed using a separate network, learned together with the auto-encoder, while the context model is learned separately.

\vspace{-0.2cm}
\section{Conclusions}
\vspace{-0.1cm}
\label{sec:conclusions}
In this paper, we proposed the first method for learning a lossy image compression auto-encoder concurrently with a lightweight context model by incorporating it into an entropy loss for the optimization of the auto-encoder, leading to performance competitive with the current state-of-the-art in deep image compression \cite{rippel17a}.

Future works could explore heavier and more powerful context models, as those employed in \cite{pixelrnn,pixelcnn}. 
This could further improve compression performance and allow for sampling of natural images in a ``lossy'' manner, by sampling $\hat{\z}$ according to the context model and then decoding.

\paragraph{Acknowledgements}
This work was supported by ETH Z\"urich and by NVIDIA through a GPU grant.

{\small
\bibliographystyle{ieee}
\bibliography{egbib.bib}
}

\newpage

\appendix
\section*{Conditional Probability Models for Deep \\Image Compression -- Suppl. Material}

\begin{figure}[!ht]
\centering
\vspace{-0.4cm}
\includegraphics[width=\linewidth]{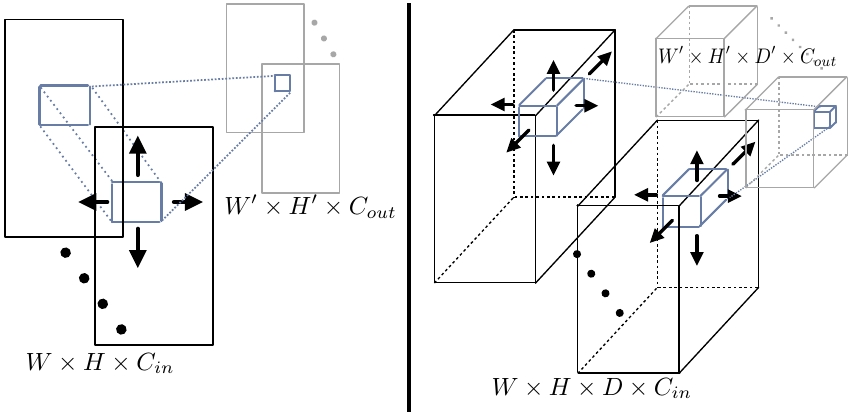}
\vspace{-0.3cm}
\caption{2D vs.\ 3D CNNs}
\label{fig:conv_2dvs3d}
\vspace{-0.4cm}
\end{figure}

\paragraph{3D probability classifier}

As mentioned in Section~\ref{sec:entropyest}, we rely on masked 3D convolutions to enforce the causality constraint in our probability classifier $P$. In a 2D-CNN, standard 2D convolutions are used in filter banks, as shown in Fig.~\ref{fig:conv_2dvs3d} on the left: A $W \times H \times C_\text{in}$-dimensional tensor is mapped to a $W' \times H' \times C_\text{out}$-dimensional tensor using $C_\text{out}$ banks of $C_\text{in}$ 2D filters, i.e., filters can be represented as $f_W \times f_H \times C_\text{in} \times C_\text{out}$-dimensional tensors. Note that all $C_\text{in}$ channels are used together, which violates causality: When we encode, we proceed channel by channel.

Using 3D convolutions, a depth dimension $D$ is introduced. In a 3D-CNN, $W \times H \times D \times C_\text{in}$-dimensional tensors are mapped to $W' \times H' \times D' \times C_\text{out}$-dimensional tensors, with $f_W \times f_H \times f_D \times C_\text{in} \times C_\text{out}$-dimensional filters. Thus, a 3D-CNN slides over the depth dimension, as shown in Fig.~\ref{fig:conv_2dvs3d} on the right. We use such a 3D-CNN for $P$, where we use as input our $W \times H \times K$-dimensional feature map $\hat \z$, using $D=K, C_\text{in} = 1$ for the first layer.

\begin{figure}[!h]
\centering
\includegraphics[width=0.9\linewidth]{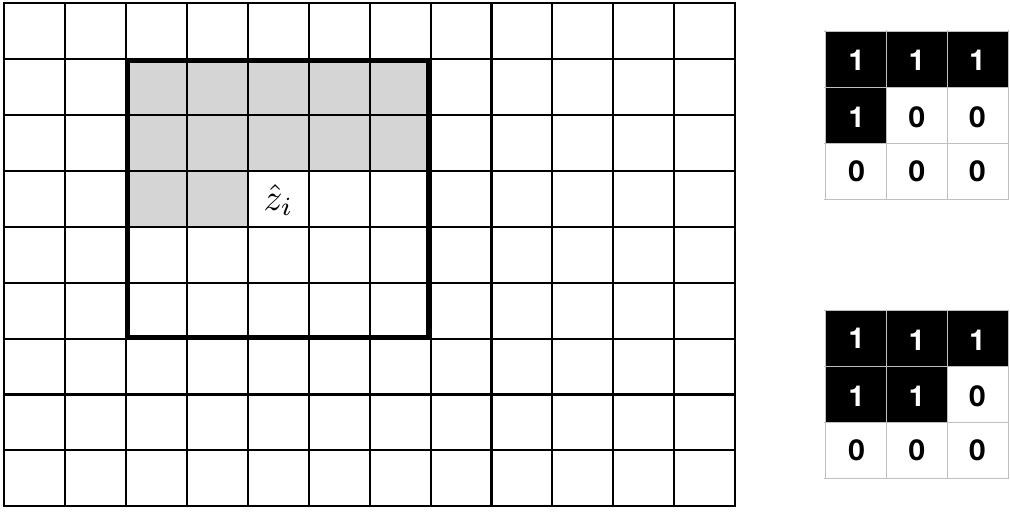}
\caption{Left shows a grid of symbols $\hat z_i$, where the black square denotes some context and the gray cells denote symbols which where previously encoded. Right shows masks.}
\label{fig:masks}
\end{figure}

To explain how we mask the filters in $P$, consider the 2D case in Fig.~\ref{fig:masks}. We want to encode all values $\hat z_i$ by iterating in raster scan order and by computing $p(\hat z_i|\hat z_{i-1}, \dots, \hat z_1)$. We simplify this by instead of relying on all previously encoded symbols, we use some $c \times c$-context around $\hat z_i$ (black square in Fig.~\ref{fig:masks}). To satisfy the causality constraint, this context may only contain values above $\hat z_i$ or in the same row to the left of $\hat z_i$ (gray cells). By using the filter shown in Fig.~\ref{fig:masks} in the top right for the first layer of a CNN and the filter shown in Fig.~\ref{fig:masks} in the bottom right for subsequent filters, we can build a 2D-CNN with a $c \times c$ receptive field that forms such a context. We build our 3D-CNN $P$ by generalizing this idea to 3D, where we construct the mask for the filter of the first layer as shown in pseudo-code Algorithm~\ref{alg:maskmaker}. The mask for the subsequent layers is constructed analoguously by replacing ``$<$'' in line~\ref{alg:maskmaker:lineLT} with ``$\leq$''. We use filter size $f_W = f_H = f_D = 3$.

\setcounter{algorithm}{-1}
\begin{algorithm}
\caption{My algorithm}\label{euclid}
\begin{algorithmic}[1]
    \State $\textit{central\_idx} \gets \ceil{(f_W \cdot f_H \cdot f_D) / 2}$
    \State $\textit{current\_idx} \gets 1$
    \State $\textit{mask} \gets f_W \times f_H \times f_D \text{-dimensional matrix of zeros}$
    \For {$d \in \{1, \dots, f_D\}$}
        \For {$h \in \{1, \dots, f_H\}$}
            \For {$w \in \{1, \dots, f_W\}$}
                \If {$\textit{current\_idx} < \textit{central\_idx}$} \label{alg:maskmaker:lineLT}
                    \State $mask(w, h, d) = 1$
                \Else
                    \State $mask(w, h, d) = 0$
                \EndIf
                \State $\textit{current\_idx} \gets \textit{current\_idx} + 1$
            \EndFor
        \EndFor
    \EndFor
\end{algorithmic}
\caption{Constructing 3D Masks}
\label{alg:maskmaker}
\end{algorithm}

With this approach, we obtain a 3D-CNN $P$ which operates on $f_H \times f_W \times f_D$-dimensional blocks. We can use $P$ to encode $\hat \z$ by iterating over $\hat \z$ in such blocks, exhausting first axis $w$, then axis $h$, and finally axis $d$ (like in Algorithm~\ref{alg:maskmaker}). For each such block, $P$ yields the probability distribution of the central symbol given the symbols in the block. Due to the construction of the masks, this probability distribution only depends on previously encoded symbols.

\paragraph{Multiple compression rates}

It is quite straightforward to obtain multiple operating points in a single network with our framework: We can simply share the network but use multiple importance maps. 
We did a simple experiment where we trained an autoencoder with 5 different importance maps. In each iteration, a random importance map $i$ was picked, and the target entropy was set to $i/5 \cdot t$. While not tuned for performance, this already yielded a model competitive with BPG. The following shows the output of the model for $i=1,3,5$ (from left to right):
{\noindent \includegraphics[width=1\linewidth]{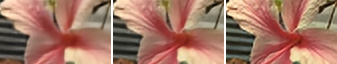}}

\begin{figure}[h]
\vspace{-1.9ex}
\begin{center}
   \includegraphics[width=\linewidth]{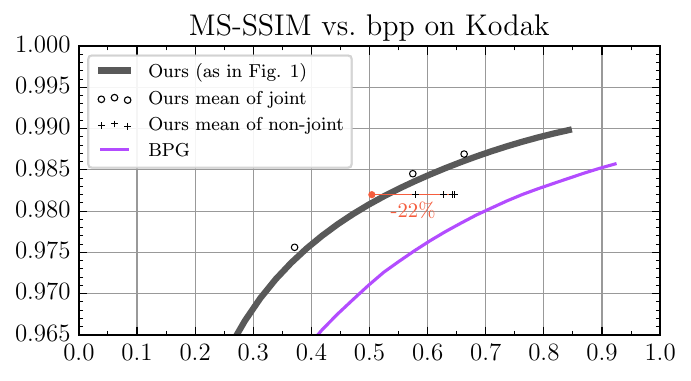}
\end{center}
\vspace{-3.5ex}
\caption{Performance on the Kodak dataset. See text.}
\label{fig:kodak}
\vspace{-1.9ex}
\end{figure}
\paragraph{On the benefit of 3DCNN and joint training}
We note that the points from Table~\ref{tab:ablationcm} (where we trained different entropy models non-jointly as a post-training step) are not directly comparable with the curve in Fig.~\ref{fig:mean_kodak}. This is because these points are obtained by taking the mean of the MS-SSIM and bpp values over the Kodak images for a single model.
In contrast, the curve in Fig.~\ref{fig:mean_kodak} is obtained by following the approach of~\cite{rippel17a}, constructing a MS-SSIM vs.\ bpp curve per-image via interpolation (see \emph{Comparison} in Section~\ref{sec:experiments}).
In Fig.~\ref{fig:kodak}, we show the black curve from Fig.~\ref{fig:mean_kodak}, as well as the mean (MS-SSIM, bpp) points achieved by the underlying models ($\circ$).
We also show the points from Tab.~\ref{tab:ablationcm} ($+$).
We can see that our masked 3DCNN with joint training gives a significant improvement over the separately trained 3DCNN, i.e., a 22\% reduction in bpp when comparing mean points (the red point is estimated).

\paragraph{Non-realistic images}

\begin{figure}
    \centering
    \includegraphics[width=0.8\linewidth]{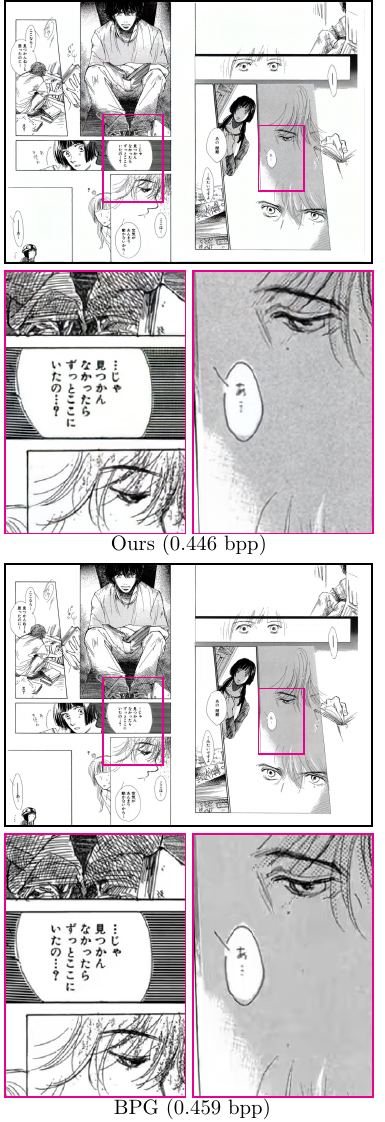}
    \vspace{-1ex}
    \caption{Comparison on a non-realistic image. See text.}
    \vspace{-2ex}
    \label{fig:manga}
\end{figure}

In Fig.~\ref{fig:manga}, we compare our approach to BPG on an image from the Manga109\footnote{\url{http://www.manga109.org/}} dataset. We can see that our approach preserves text well enough to still be legible, but it is not as crip as BPG (left zoom). On the other hand, our approach manages to preserve the fine texture on the face better than BPG (right zoom).

\paragraph{Visual examples}

The following pages show the first four images of each of our validation sets compressed to low bitrate, together with outputs from BPG, JPEG2000 and JPEG compressed to similar bitrates. We ignored all header information for all considered methods when computing the bitrate (here and throughout the paper).
We note that the only header our approach requires is the size of the image and an identifier, e.g., $\beta$, specifying the model.

\begin{figure*}[!h]
\captionsetup{width=0.807\linewidth}
\centering
\setlength{\tabcolsep}{1pt}
\begin{tabular}{lr}
    
\includegraphics[width=0.4\textwidth]{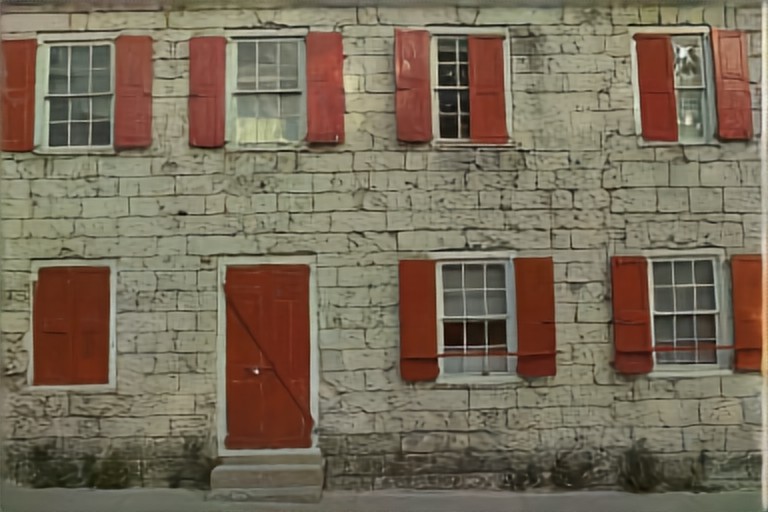}&
\includegraphics[width=0.4\textwidth]{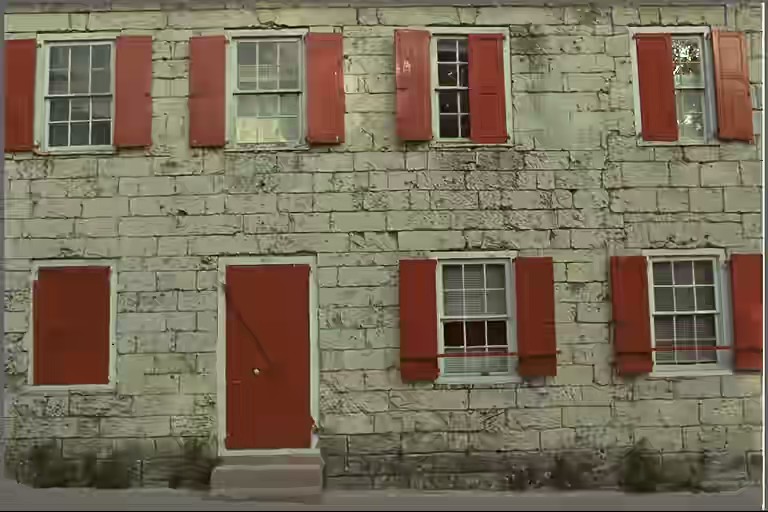}\\[-0.5ex]
\textbf{Ours} 0.239 bpp & 0.246 bpp \textbf{BPG} \\
\includegraphics[width=0.4\textwidth]{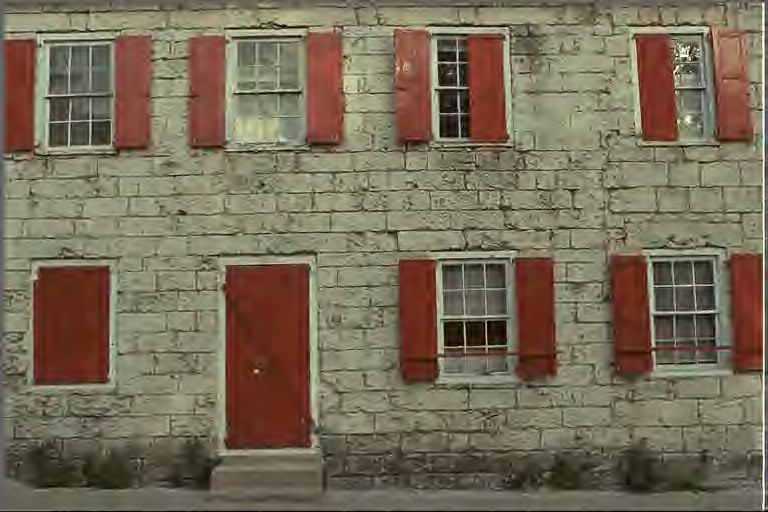}&
\includegraphics[width=0.4\textwidth]{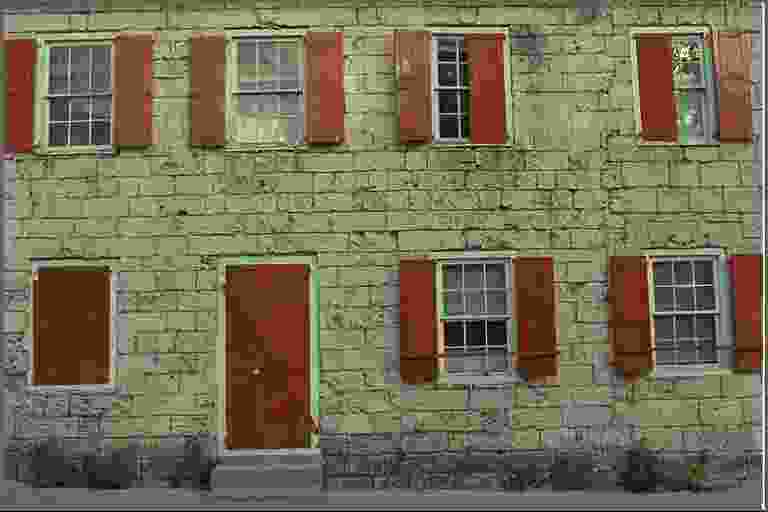}\\
\textbf{JPEG 2000} 0.242 bpp & 0.259 bpp \textbf{JPEG}
\\[0.5cm]
    
\includegraphics[width=0.4\textwidth]{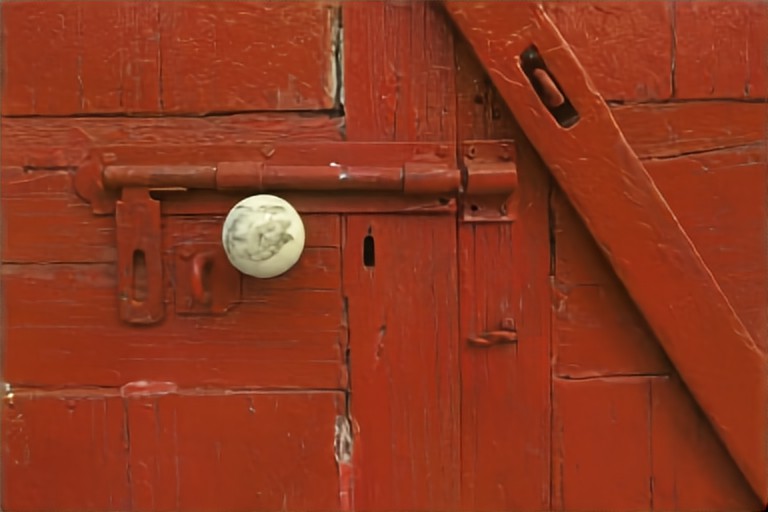}&
\includegraphics[width=0.4\textwidth]{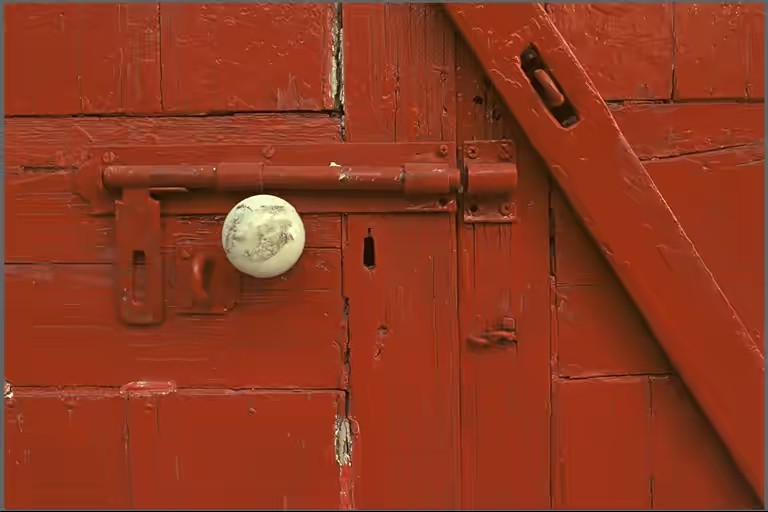}\\[-0.5ex]
\textbf{Ours} 0.203 bpp & 0.201 bpp \textbf{BPG} \\
\includegraphics[width=0.4\textwidth]{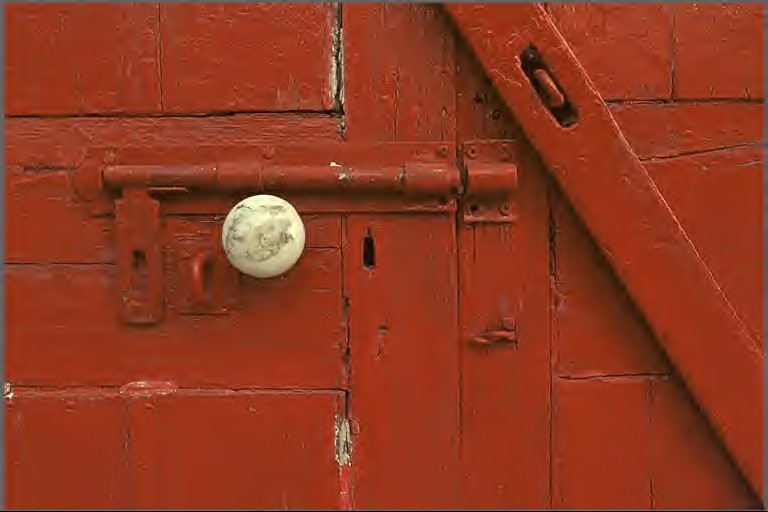}&
\includegraphics[width=0.4\textwidth]{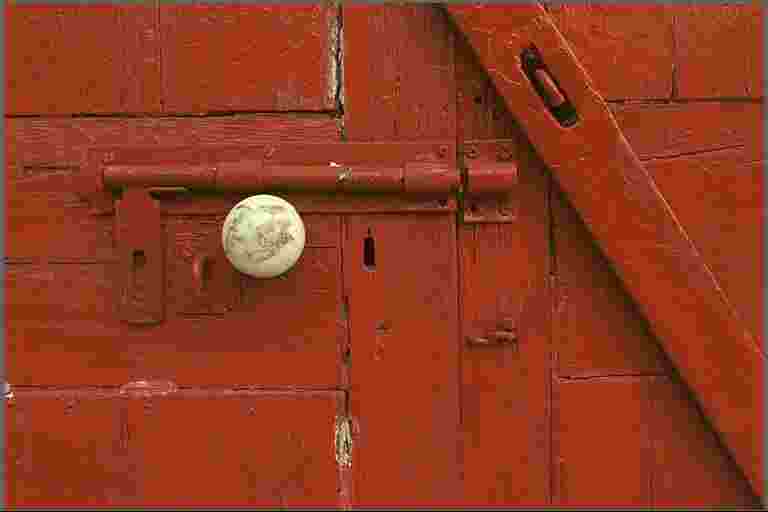}\\
\textbf{JPEG 2000} 0.197 bpp & 0.205 bpp \textbf{JPEG}

\end{tabular}
\vspace{-0.2cm}
\caption{\label{fig:vis_ex_Kodak_first}Our approach vs.\ BPG, JPEG and JPEG 2000 on the first and second image of the Kodak dataset, along with bit rate.}
\end{figure*}

\begin{figure*}[!h]
\captionsetup{width=0.807\linewidth}
\centering
\setlength{\tabcolsep}{1pt}
\begin{tabular}{lr}
    
\includegraphics[width=0.4\textwidth]{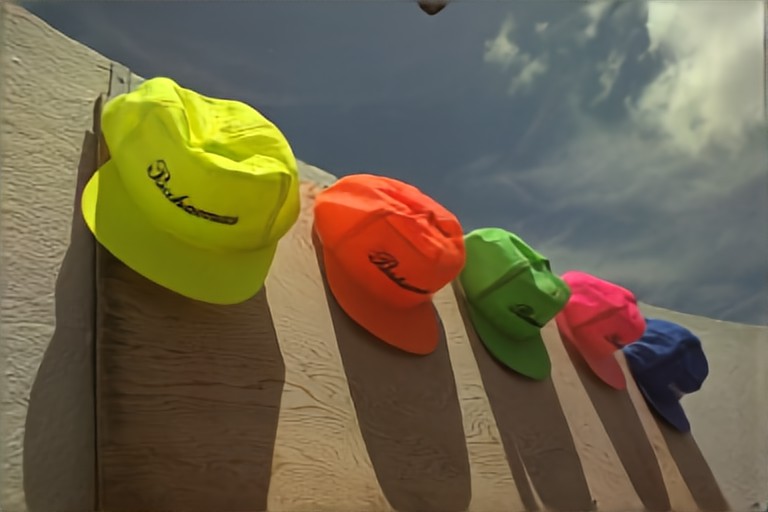}&
\includegraphics[width=0.4\textwidth]{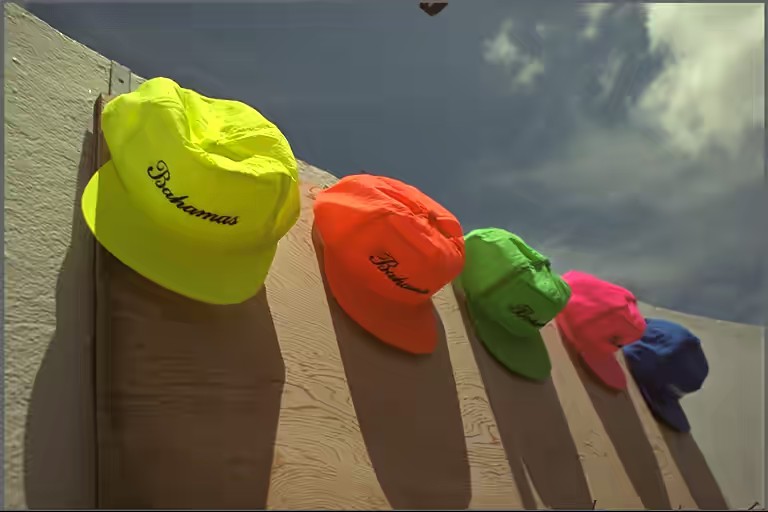}\\[-0.5ex]
\textbf{Ours} 0.165 bpp & 0.164 bpp \textbf{BPG} \\
\includegraphics[width=0.4\textwidth]{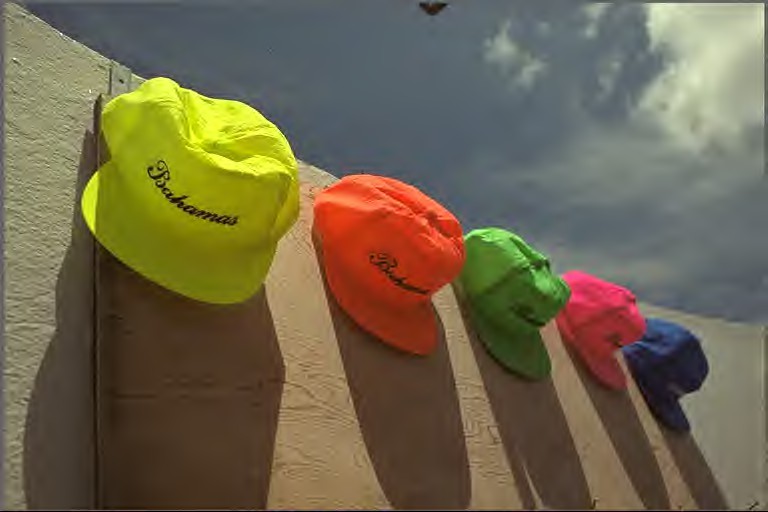}&
\includegraphics[width=0.4\textwidth]{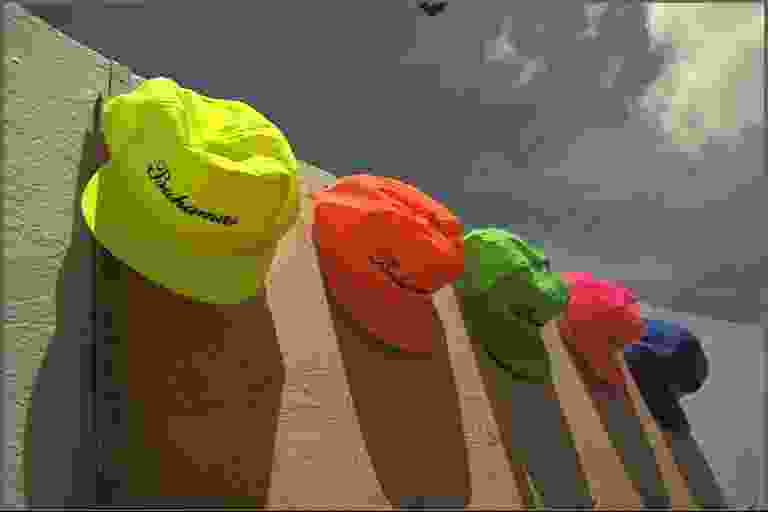}\\
\textbf{JPEG 2000} 0.166 bpp & 0.166 bpp \textbf{JPEG}
\\[0.5cm]
    
\includegraphics[width=0.4\textwidth]{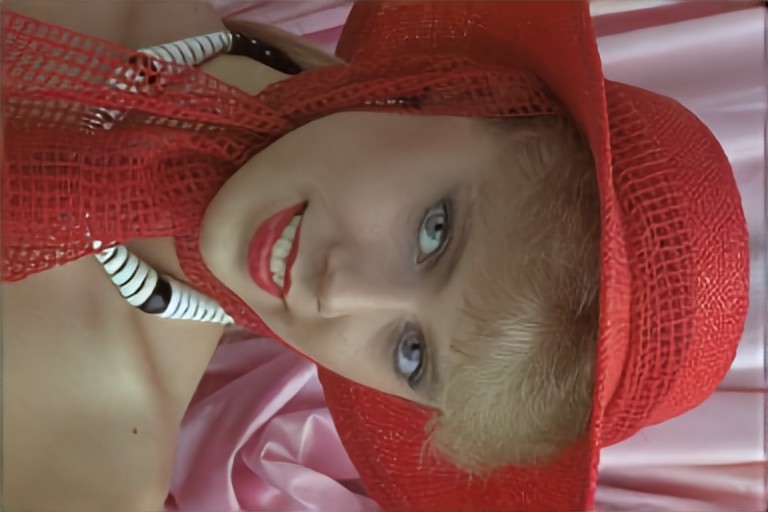}&
\includegraphics[width=0.4\textwidth]{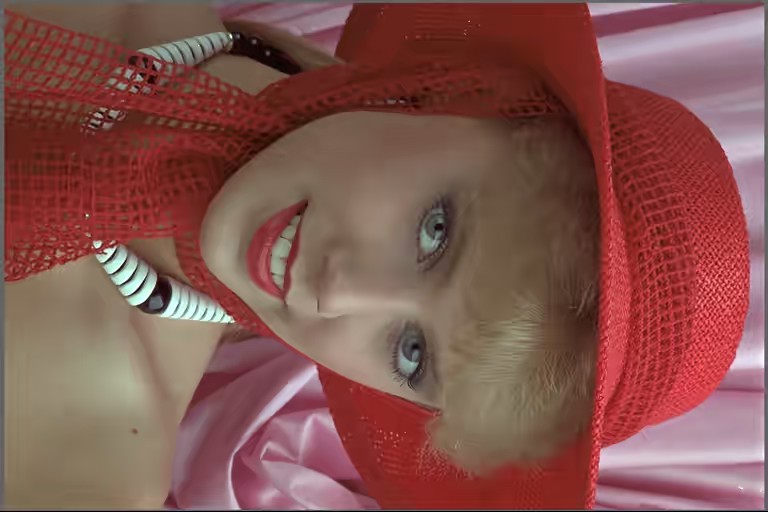}\\[-0.5ex]
\textbf{Ours} 0.193 bpp & 0.209 bpp \textbf{BPG} \\
\includegraphics[width=0.4\textwidth]{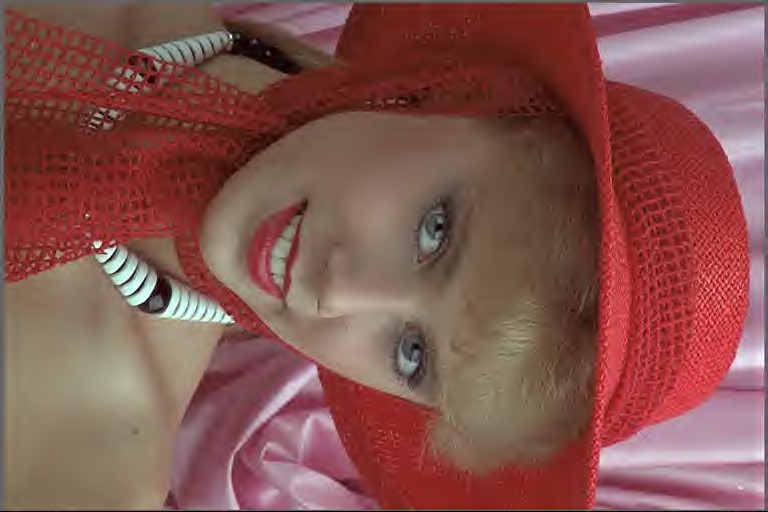}&
\includegraphics[width=0.4\textwidth]{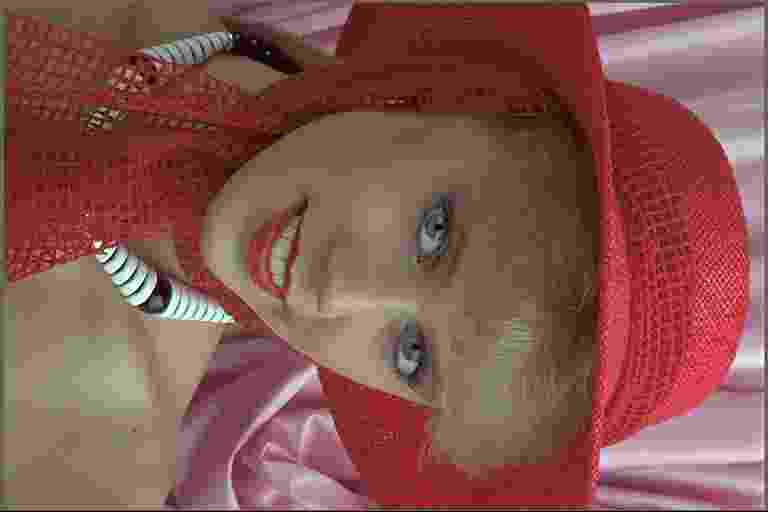}\\
\textbf{JPEG 2000} 0.194 bpp & 0.203 bpp \textbf{JPEG}

\end{tabular}
\vspace{-0.2cm}
\caption{\label{fig:vis_ex_Kodak_third}Our approach vs.\ BPG, JPEG and JPEG 2000 on the third and fourth image of the Kodak dataset, along with bit rate.}
\end{figure*}

\begin{figure*}[!h]
\captionsetup{width=0.807\linewidth}
\centering
\setlength{\tabcolsep}{1pt}
\begin{tabular}{lr}
    
\includegraphics[width=0.4\textwidth]{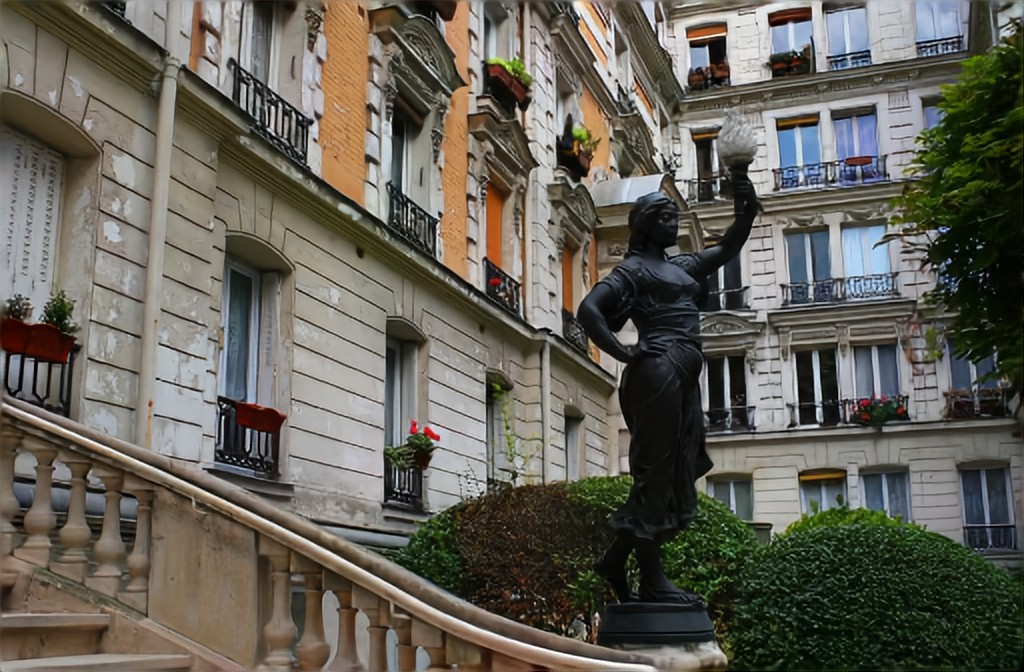}&
\includegraphics[width=0.4\textwidth]{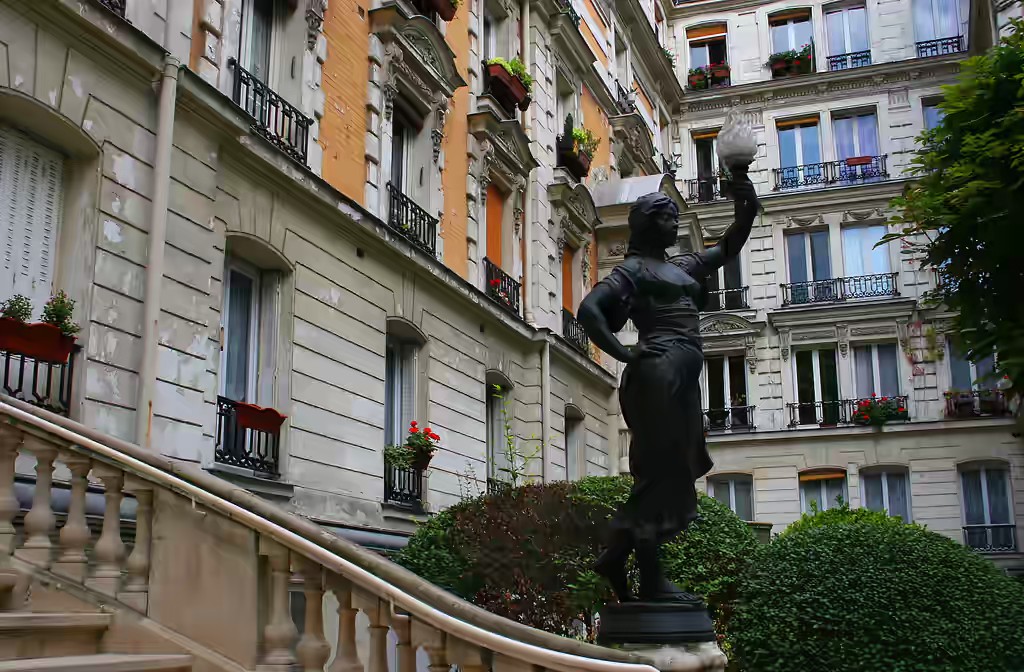}\\[-0.5ex]
\textbf{Ours} 0.435 bpp & 0.479 bpp \textbf{BPG} \\
\includegraphics[width=0.4\textwidth]{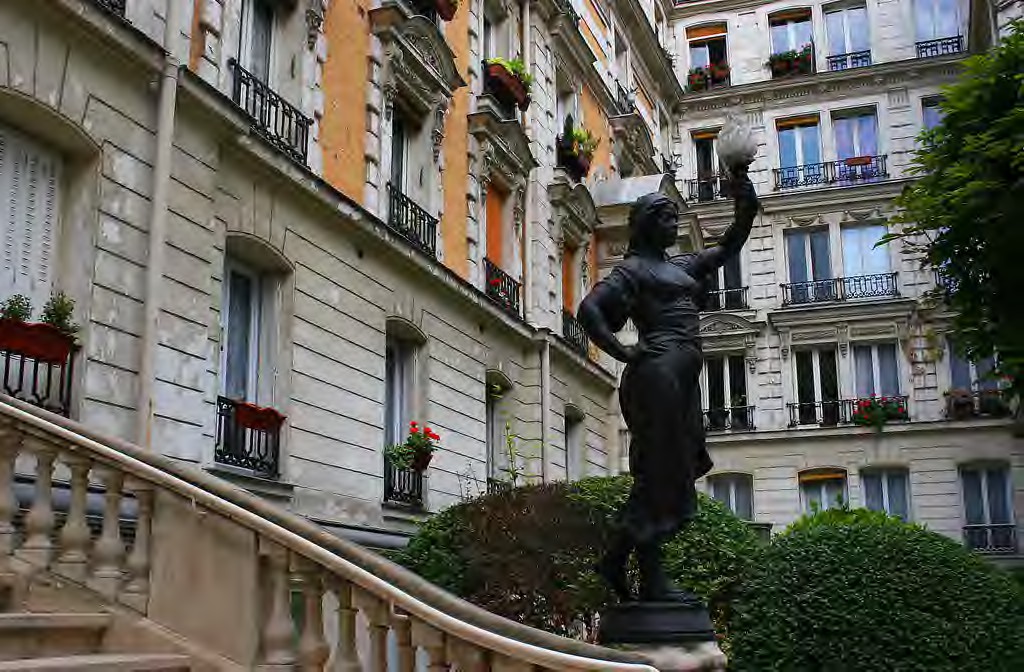}&
\includegraphics[width=0.4\textwidth]{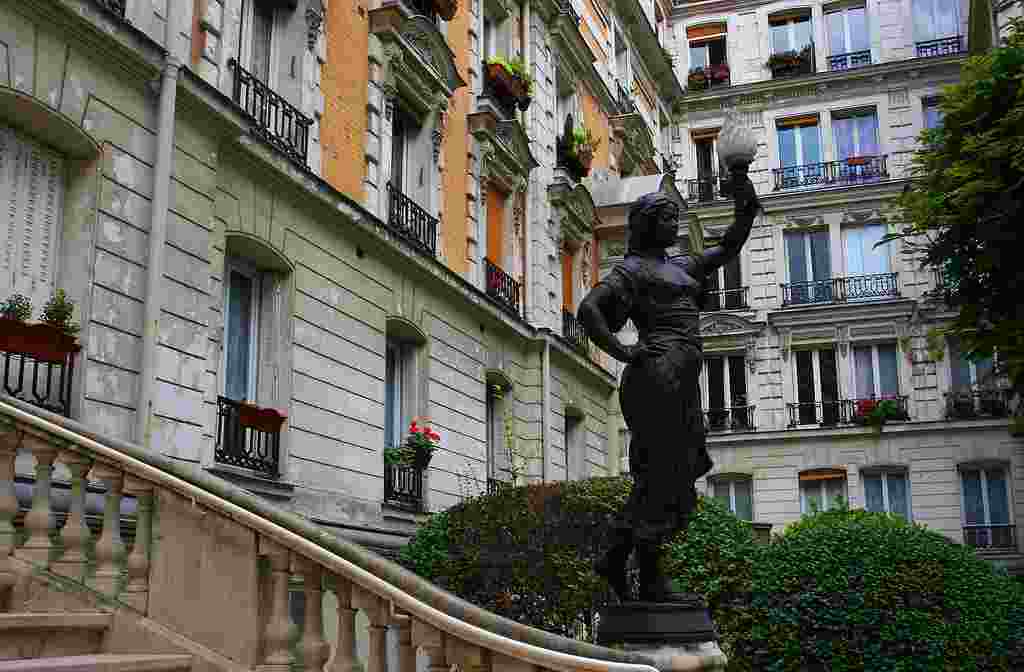}\\
\textbf{JPEG 2000} 0.437 bpp & 0.445 bpp \textbf{JPEG}
\\[0.5cm]
    
\includegraphics[width=0.4\textwidth]{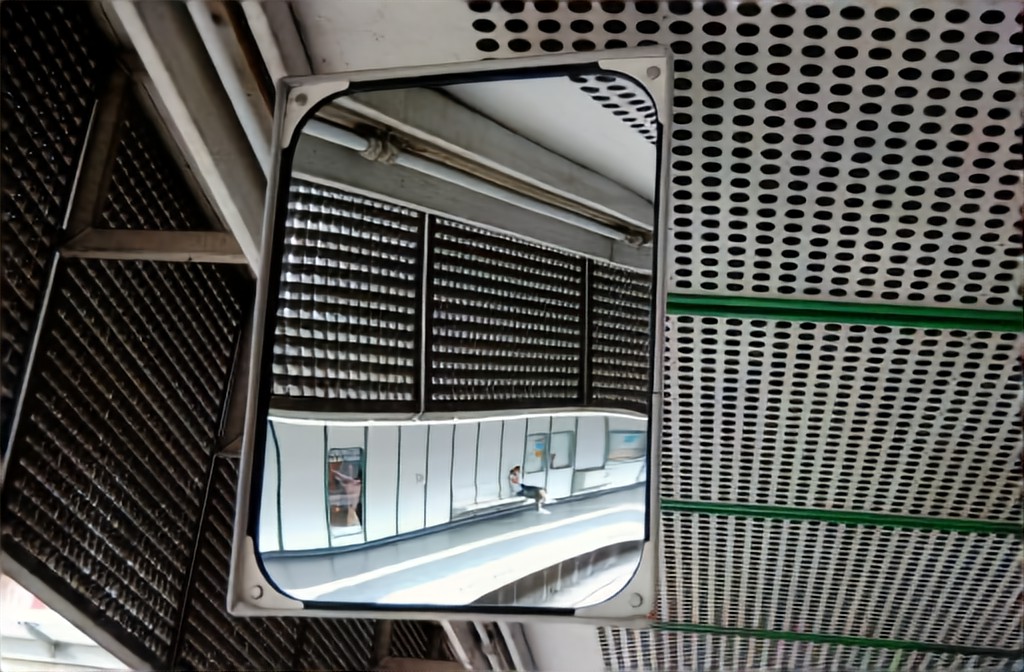}&
\includegraphics[width=0.4\textwidth]{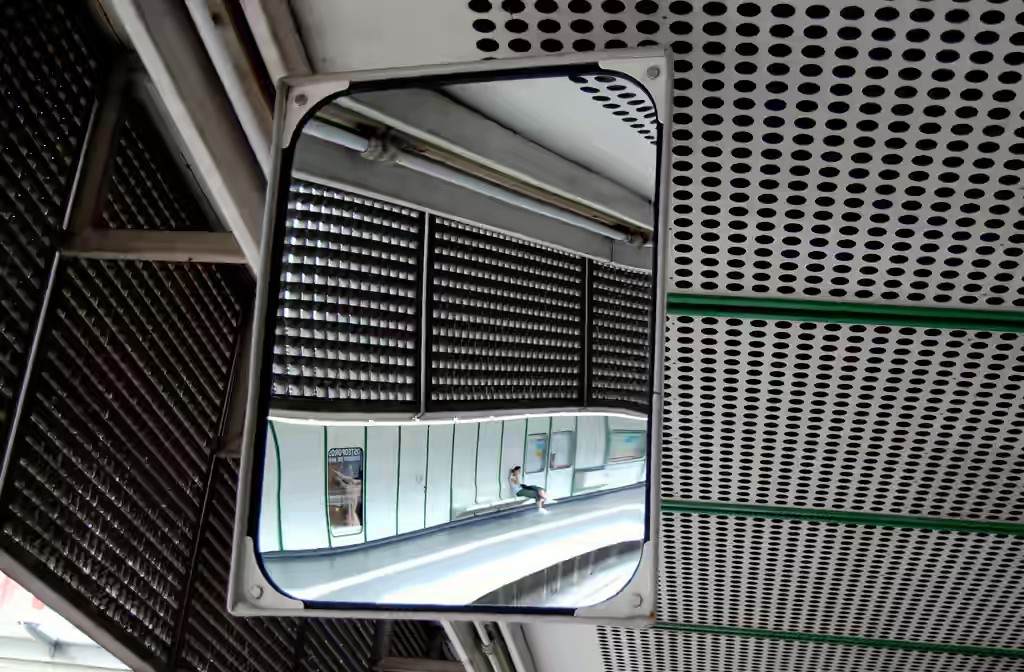}\\[-0.5ex]
\textbf{Ours} 0.345 bpp & 0.377 bpp \textbf{BPG} \\
\includegraphics[width=0.4\textwidth]{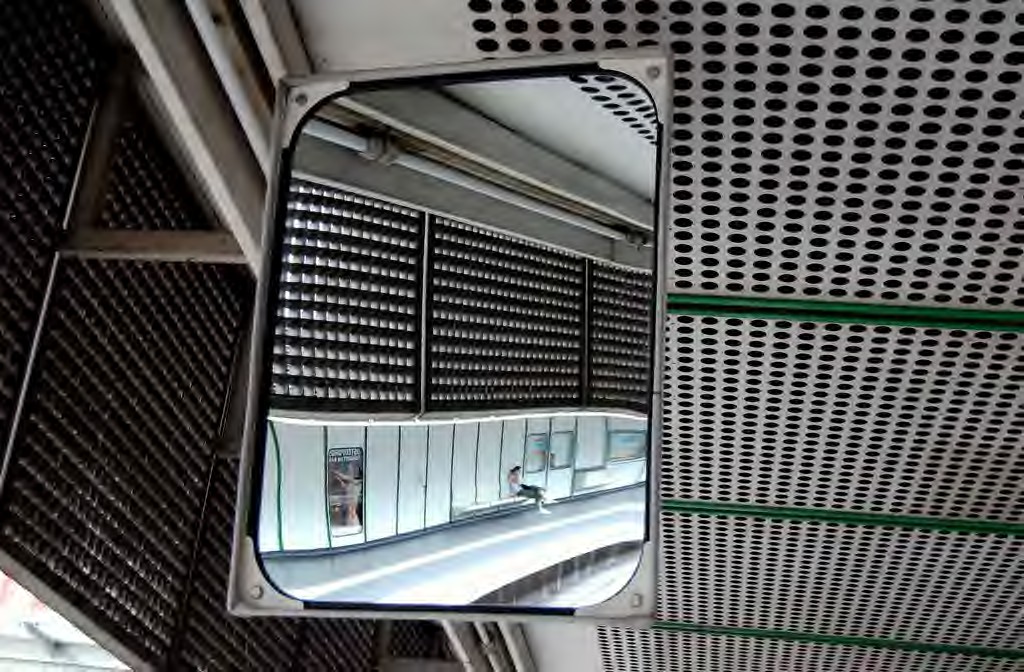}&
\includegraphics[width=0.4\textwidth]{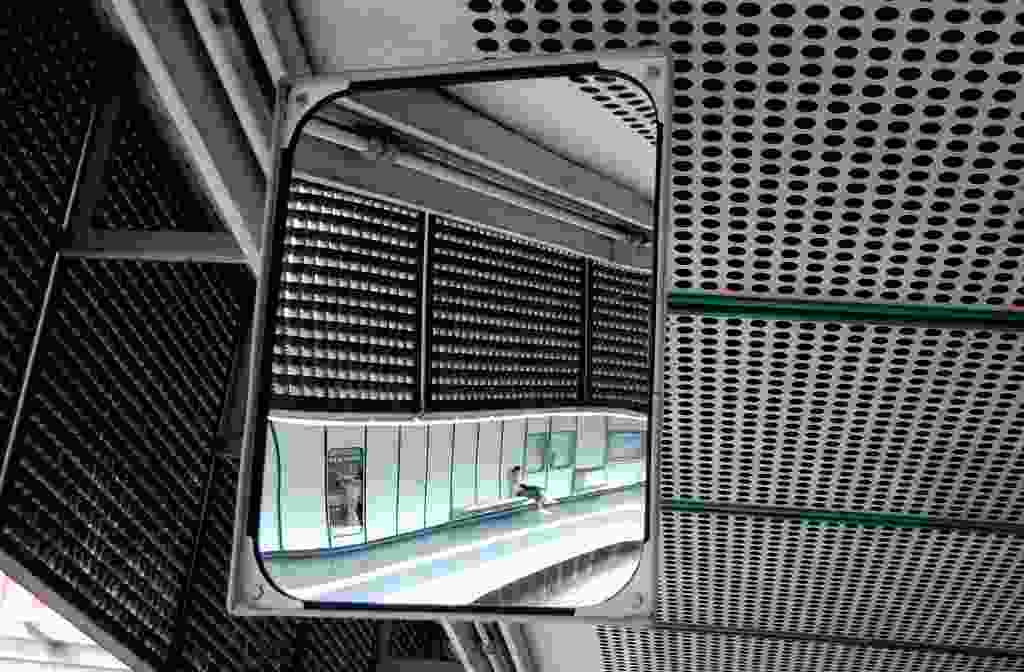}\\
\textbf{JPEG 2000} 0.349 bpp & 0.357 bpp \textbf{JPEG}

\end{tabular}
\vspace{-0.2cm}
\caption{\label{fig:vis_ex_Urban100_third}Our approach vs.\ BPG, JPEG and JPEG 2000 on the third and fourth image of the Urban100 dataset, along with bit rate.}
\end{figure*}

\begin{figure*}[!h]
\captionsetup{width=0.807\linewidth}
\centering
\setlength{\tabcolsep}{1pt}
\begin{tabular}{lr}
    
\includegraphics[width=0.4\textwidth]{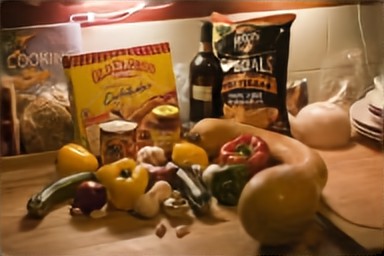}&
\includegraphics[width=0.4\textwidth]{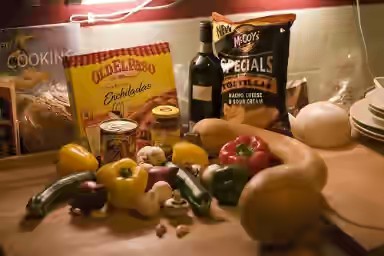}\\[-0.5ex]
\textbf{Ours} 0.355 bpp & 0.394 bpp \textbf{BPG} \\
\includegraphics[width=0.4\textwidth]{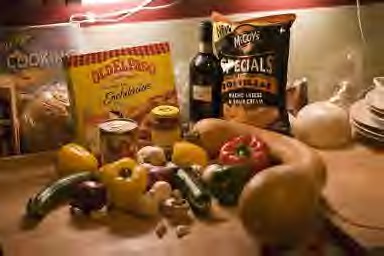}&
\includegraphics[width=0.4\textwidth]{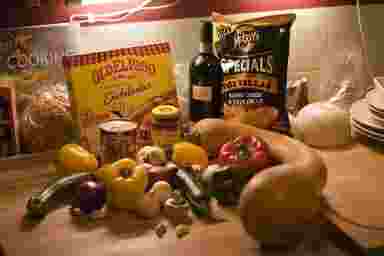}\\
\textbf{JPEG 2000} 0.349 bpp & 0.378 bpp \textbf{JPEG}
\\[0.5cm]
    
\includegraphics[width=0.4\textwidth]{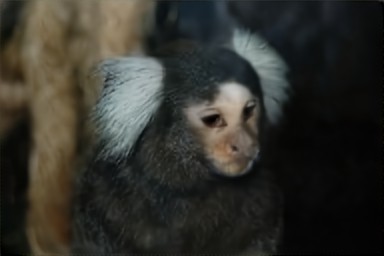}&
\includegraphics[width=0.4\textwidth]{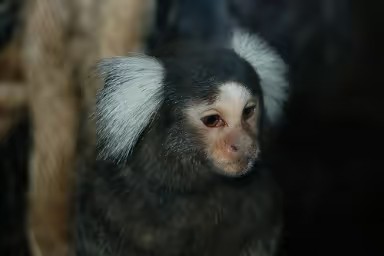}\\[-0.5ex]
\textbf{Ours} 0.263 bpp & 0.267 bpp \textbf{BPG} \\
\includegraphics[width=0.4\textwidth]{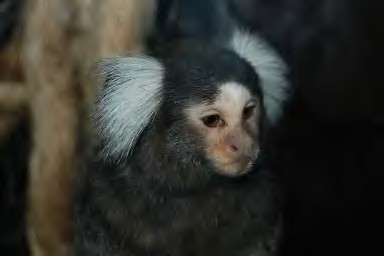}&
\includegraphics[width=0.4\textwidth]{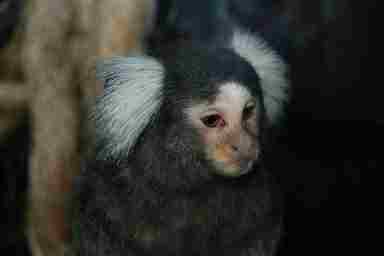}\\
\textbf{JPEG 2000} 0.254 bpp & 0.266 bpp \textbf{JPEG}

\end{tabular}
\vspace{-0.2cm}
\caption{\label{fig:vis_ex_ImageNetTest_first}Our approach vs.\ BPG, JPEG and JPEG 2000 on the first and second image of the ImageNetTest dataset, along with bit rate.}
\end{figure*}

Overall, our images look pleasant to the eye. We see cases of over-blurring in our outputs, where BPG manages to keep high frequencies due to its more local approach. An example is the fences in front of the windows in Fig.~\ref{fig:vis_ex_Urban100_third}, top, or the text in Fig.~\ref{fig:vis_ex_ImageNetTest_first}, top. On the other hand, BPG tends to discard low-contrast high frequencies where our approach keeps them in the output, like in the door in Fig.~\ref{fig:vis_ex_Kodak_first}, top, or in the hair in Fig.~\ref{fig:vis_ex_Kodak_third}, bottom. This may be explained by BPG being optimized for MSE as opposed to our approach being optimized for MS-SSIM.

JPEG looks extremely blocky for most images due to the very low bitrate.

\begin{figure*}[!h]
\captionsetup{width=0.807\linewidth}
\centering
\setlength{\tabcolsep}{1pt}
\begin{tabular}{lr}
    
\includegraphics[width=0.4\textwidth]{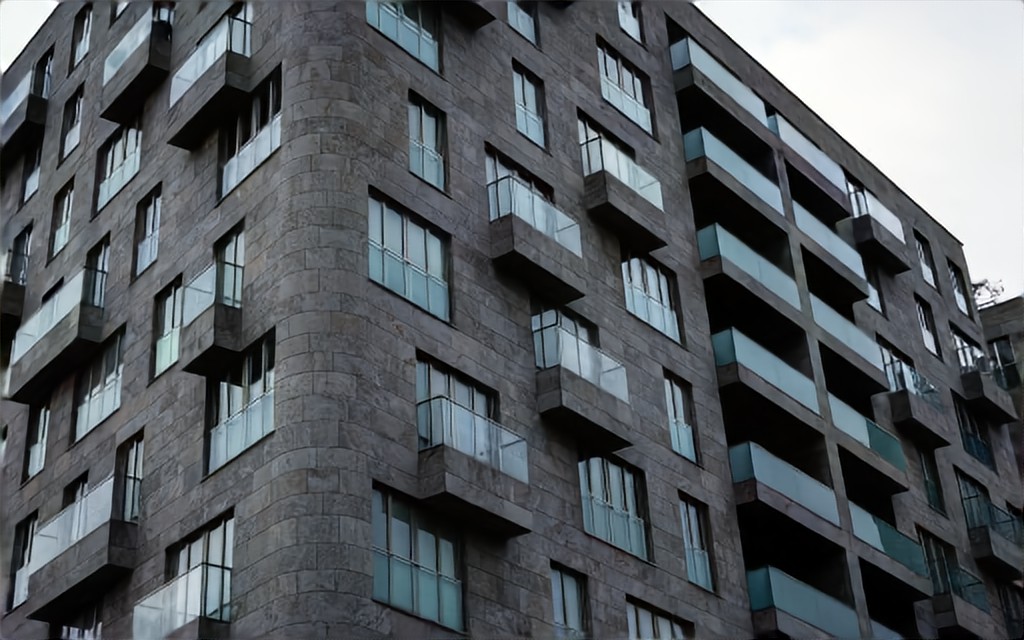}&
\includegraphics[width=0.4\textwidth]{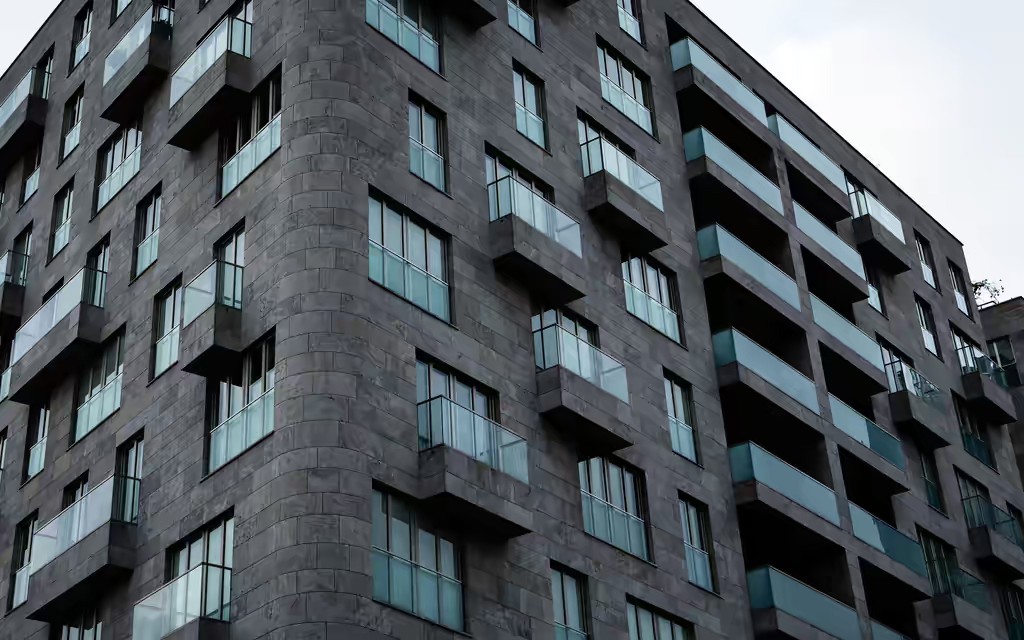}\\[-0.5ex]
\textbf{Ours} 0.385 bpp & 0.394 bpp \textbf{BPG} \\
\includegraphics[width=0.4\textwidth]{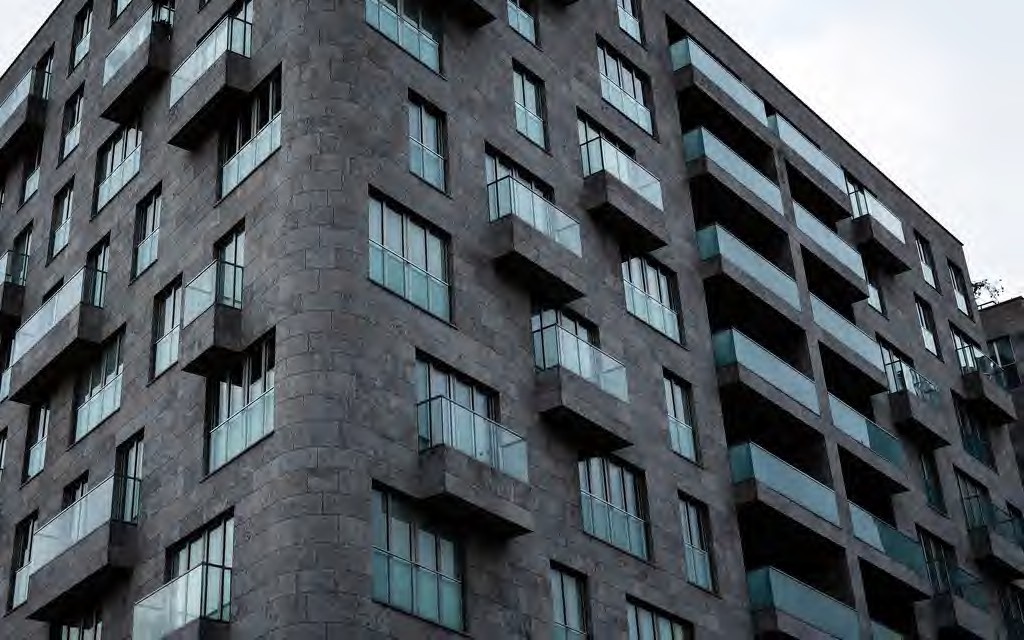}&
\includegraphics[width=0.4\textwidth]{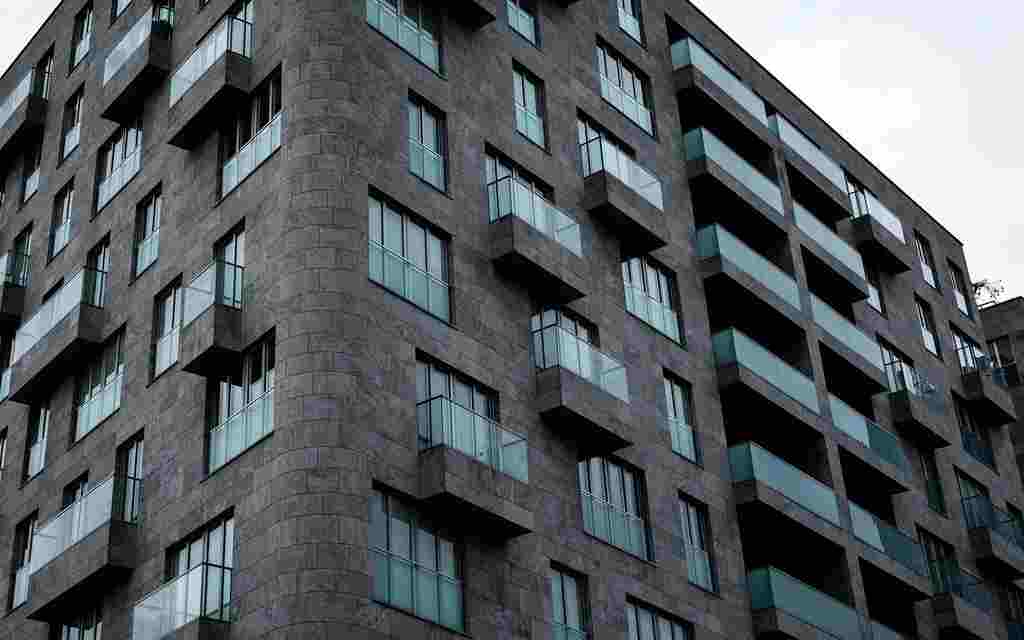}\\
\textbf{JPEG 2000} 0.377 bpp & 0.386 bpp \textbf{JPEG}
\\[0.5cm]
    
\includegraphics[width=0.4\textwidth]{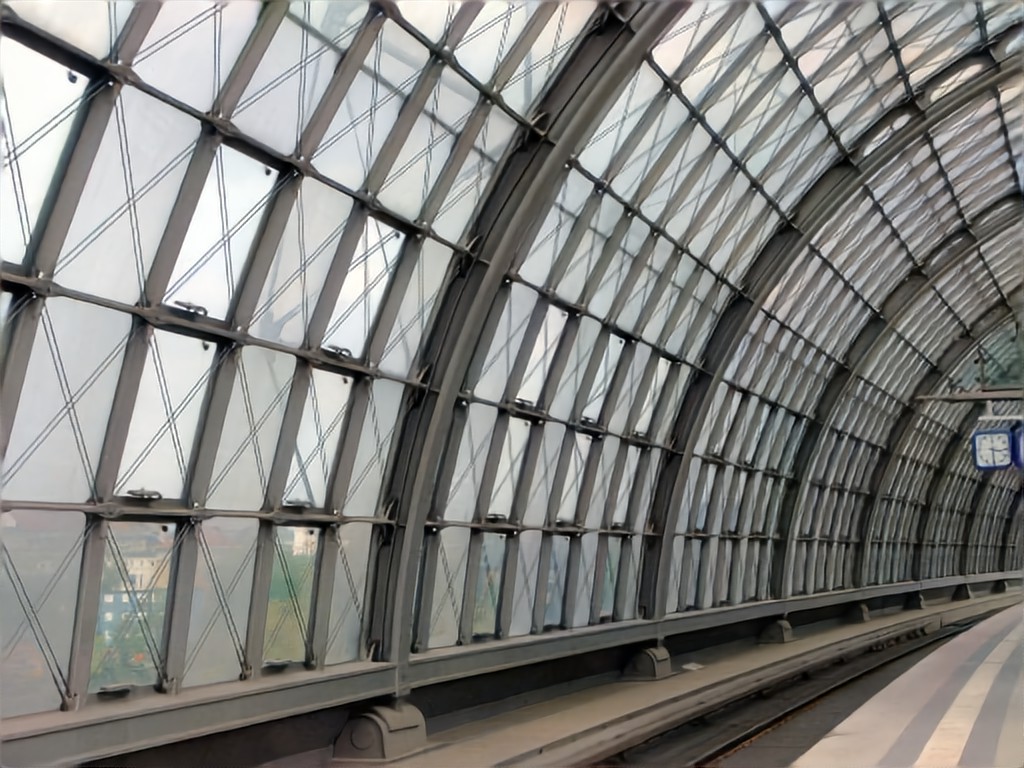}&
\includegraphics[width=0.4\textwidth]{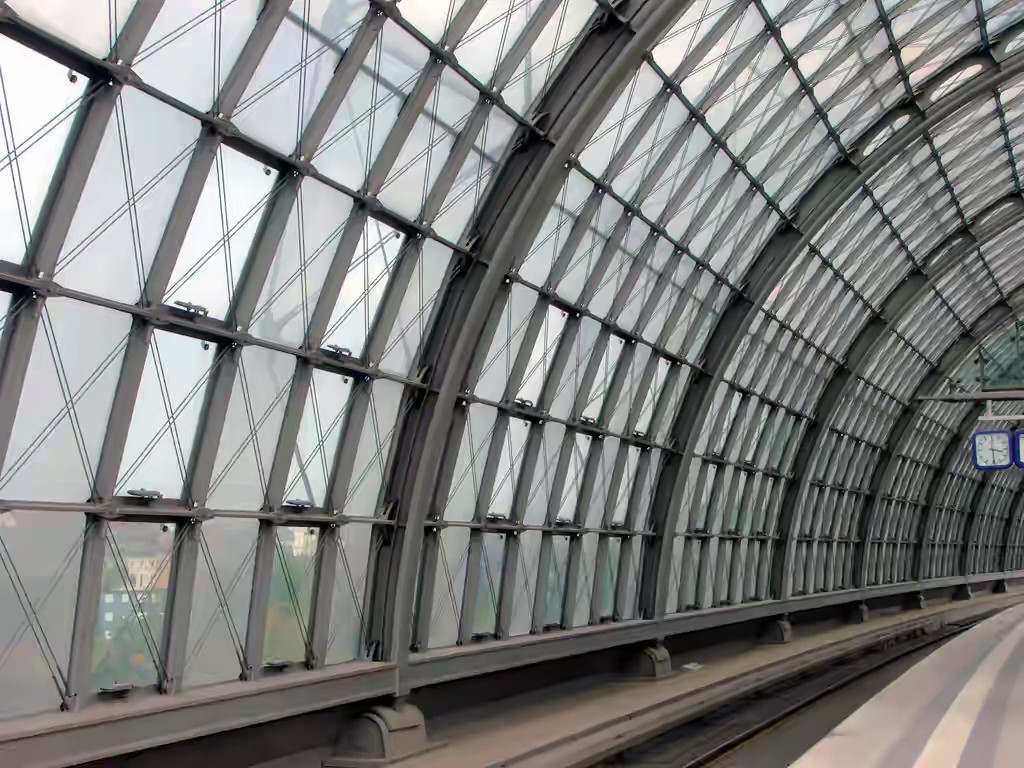}\\[-0.5ex]
\textbf{Ours} 0.365 bpp & 0.363 bpp \textbf{BPG} \\
\includegraphics[width=0.4\textwidth]{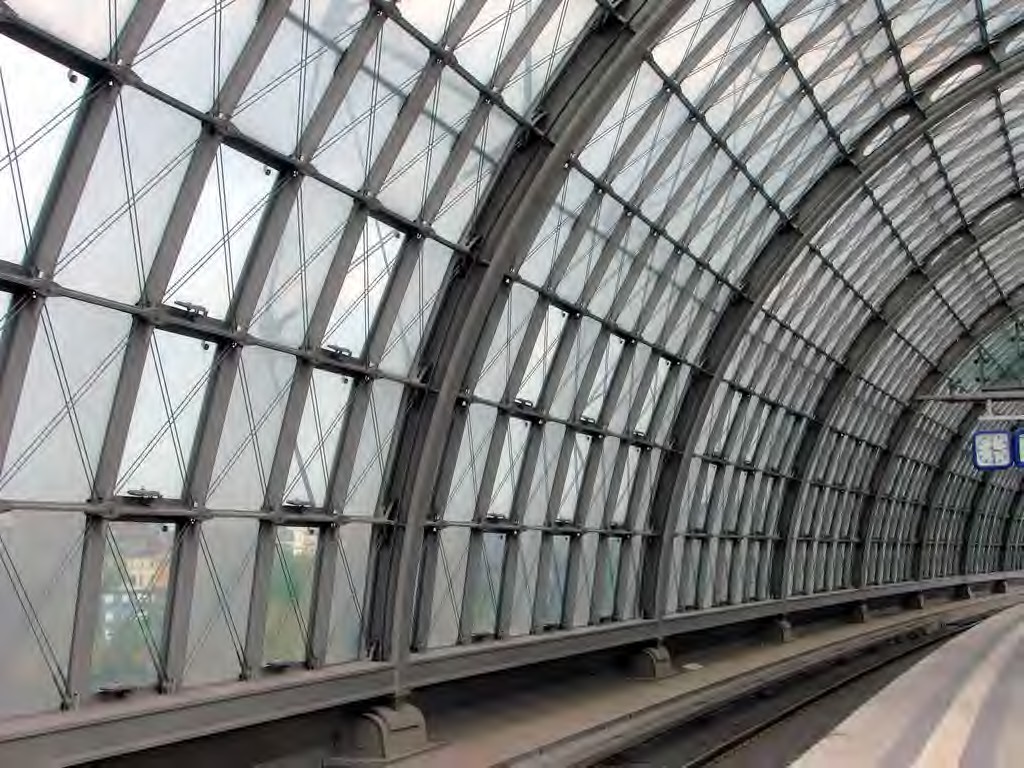}&
\includegraphics[width=0.4\textwidth]{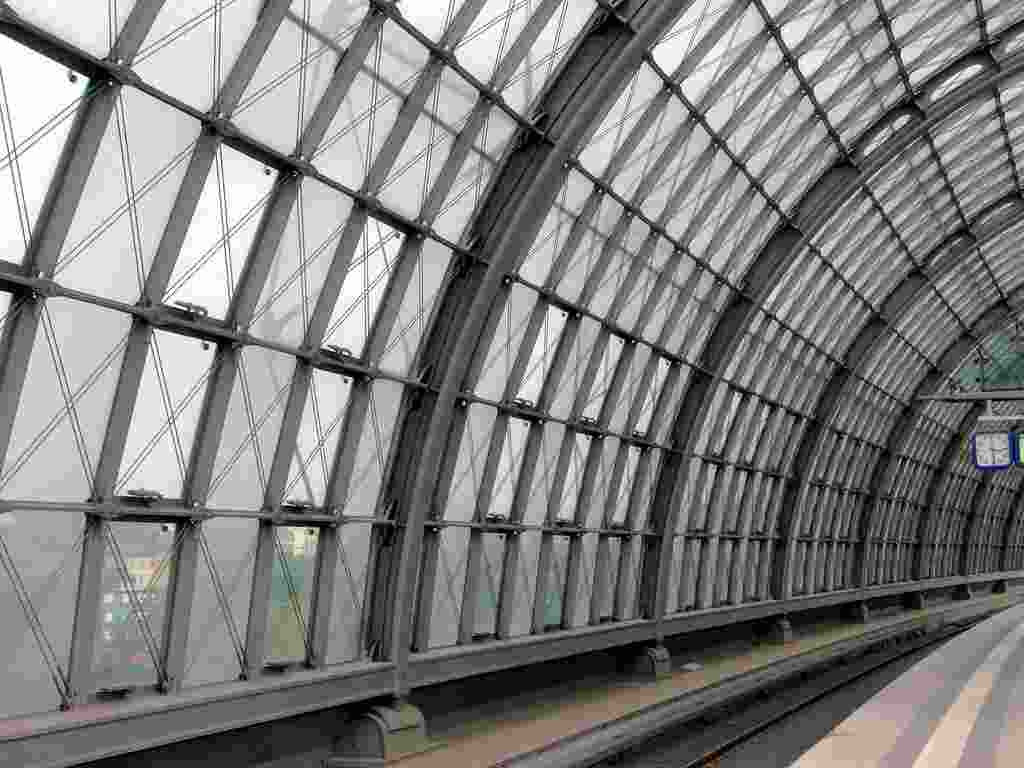}\\
\textbf{JPEG 2000} 0.363 bpp & 0.372 bpp \textbf{JPEG}

\end{tabular}
\vspace{-0.2cm}
\caption{\label{fig:vis_ex_Urban100_first}Our approach vs.\ BPG, JPEG and JPEG 2000 on the first and second image of the Urban100 dataset, along with bit rate.}
\end{figure*}

\begin{figure*}[!h]
\captionsetup{width=0.807\linewidth}
\centering
\setlength{\tabcolsep}{1pt}
\begin{tabular}{lr}
    
\includegraphics[width=0.4\textwidth]{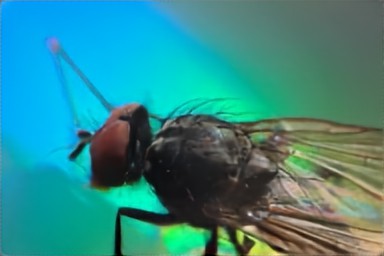}&
\includegraphics[width=0.4\textwidth]{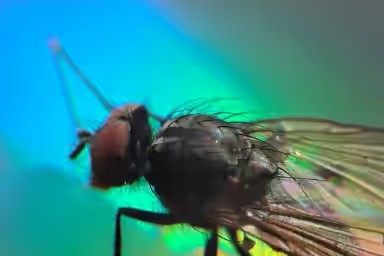}\\[-0.5ex]
\textbf{Ours} 0.284 bpp & 0.280 bpp \textbf{BPG} \\
\includegraphics[width=0.4\textwidth]{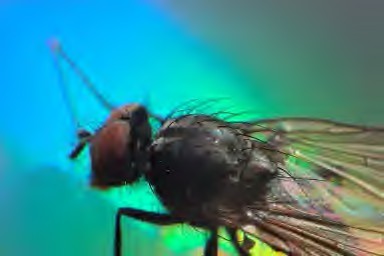}&
\includegraphics[width=0.4\textwidth]{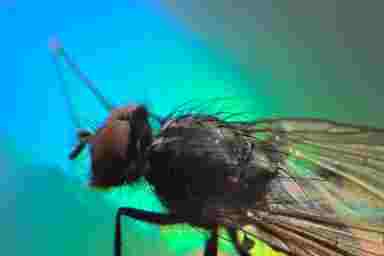}\\
\textbf{JPEG 2000} 0.287 bpp & 0.288 bpp \textbf{JPEG}
\\[0.5cm]
    
\includegraphics[width=0.4\textwidth]{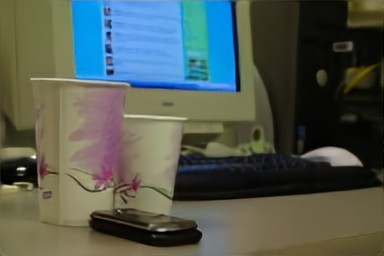}&
\includegraphics[width=0.4\textwidth]{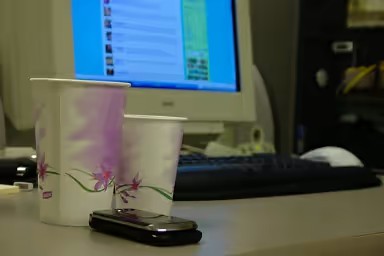}\\[-0.5ex]
\textbf{Ours} 0.247 bpp & 0.253 bpp \textbf{BPG} \\
\includegraphics[width=0.4\textwidth]{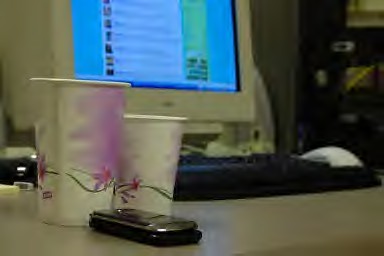}&
\includegraphics[width=0.4\textwidth]{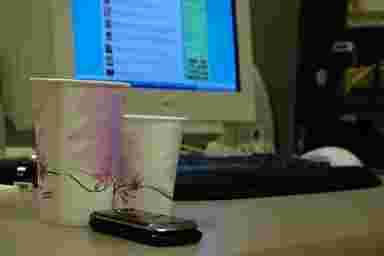}\\
\textbf{JPEG 2000} 0.243 bpp & 0.252 bpp \textbf{JPEG}

\end{tabular}
\vspace{-0.2cm}
\caption{\label{fig:vis_ex_ImageNetTest_third}Our approach vs.\ BPG, JPEG and JPEG 2000 on the third and fourth image of the ImageNetTest dataset, along with bit rate.}
\end{figure*}

\begin{figure*}[!h]
\captionsetup{width=0.807\linewidth}
\centering
\setlength{\tabcolsep}{1pt}
\begin{tabular}{lr}
    
\includegraphics[width=0.4\textwidth]{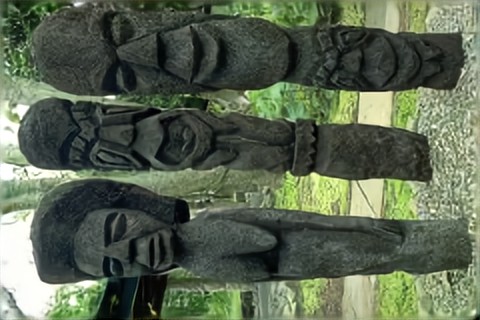}&
\includegraphics[width=0.4\textwidth]{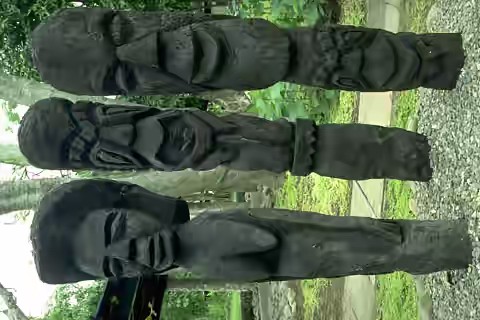}\\[-0.5ex]
\textbf{Ours} 0.494 bpp & 0.501 bpp \textbf{BPG} \\
\includegraphics[width=0.4\textwidth]{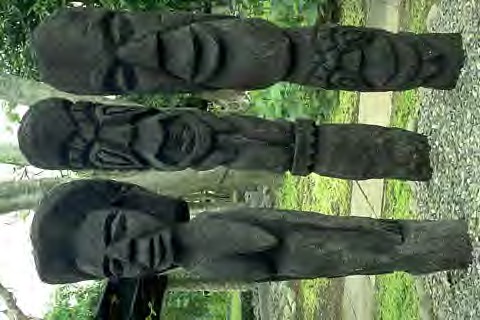}&
\includegraphics[width=0.4\textwidth]{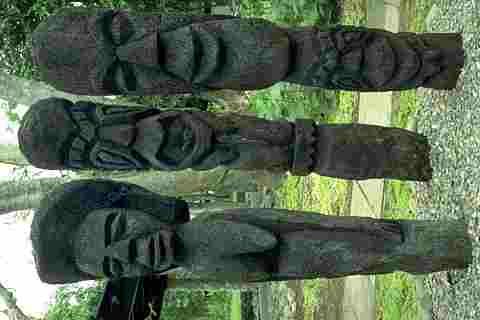}\\
\textbf{JPEG 2000} 0.490 bpp & 0.525 bpp \textbf{JPEG}
\\[0.5cm]
    
\includegraphics[width=0.4\textwidth]{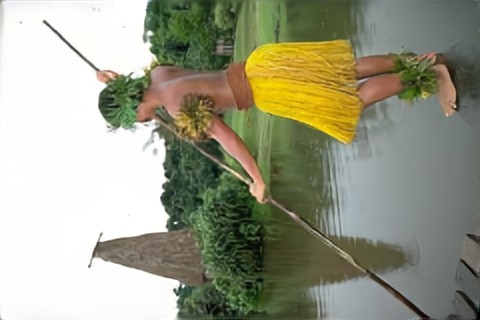}&
\includegraphics[width=0.4\textwidth]{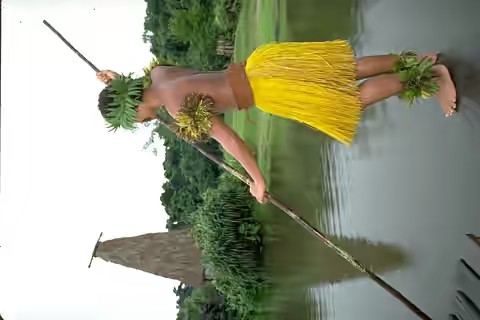}\\[-0.5ex]
\textbf{Ours} 0.298 bpp & 0.301 bpp \textbf{BPG} \\
\includegraphics[width=0.4\textwidth]{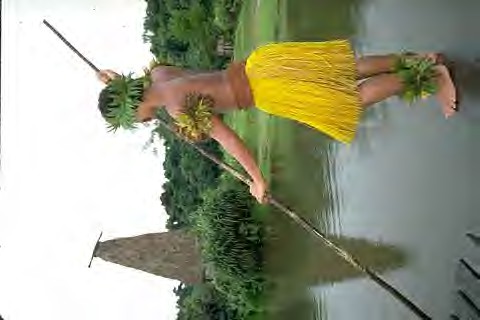}&
\includegraphics[width=0.4\textwidth]{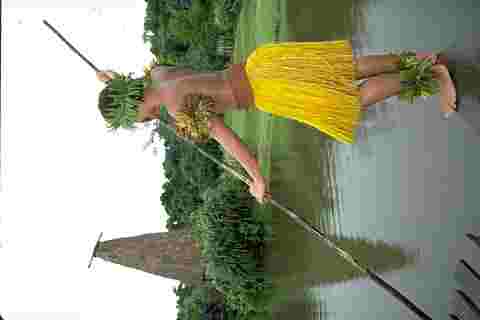}\\
\textbf{JPEG 2000} 0.293 bpp & 0.315 bpp \textbf{JPEG}

\end{tabular}
\vspace{-0.2cm}
\caption{\label{fig:vis_ex_B100_first}Our approach vs.\ BPG, JPEG and JPEG 2000 on the first and second image of the B100 dataset, along with bit rate.}
\end{figure*}

\begin{figure*}[!h]
\captionsetup{width=0.807\linewidth}
\centering
\setlength{\tabcolsep}{1pt}
\begin{tabular}{lr}
    
\includegraphics[width=0.4\textwidth]{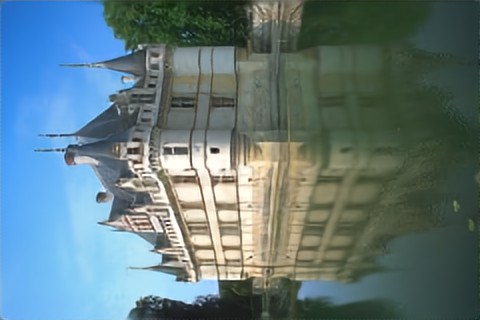}&
\includegraphics[width=0.4\textwidth]{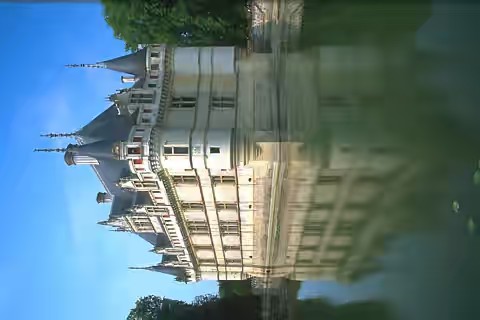}\\[-0.5ex]
\textbf{Ours} 0.315 bpp & 0.329 bpp \textbf{BPG} \\
\includegraphics[width=0.4\textwidth]{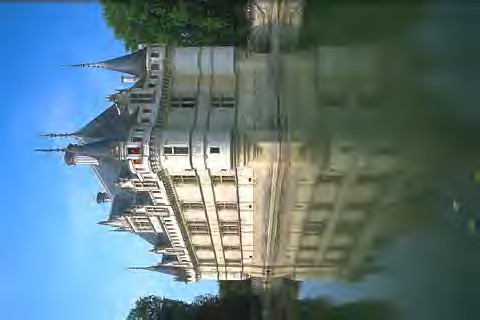}&
\includegraphics[width=0.4\textwidth]{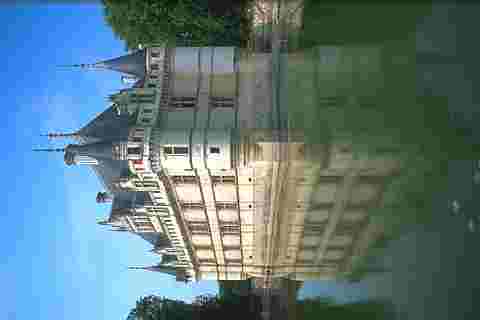}\\
\textbf{JPEG 2000} 0.311 bpp & 0.321 bpp \textbf{JPEG}
\\[0.5cm]
    
\includegraphics[width=0.4\textwidth]{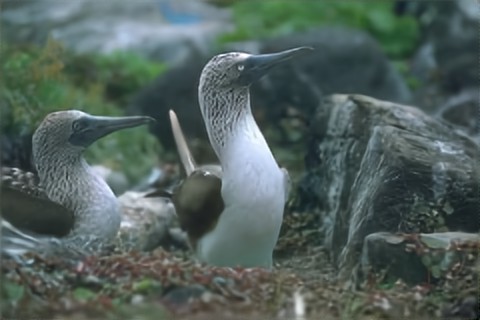}&
\includegraphics[width=0.4\textwidth]{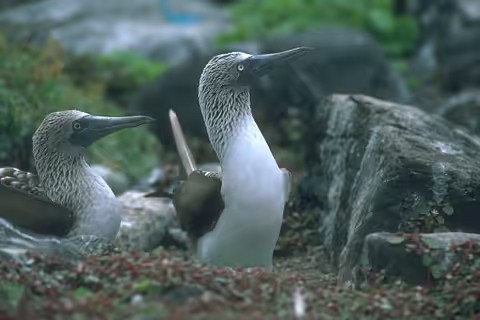}\\[-0.5ex]
\textbf{Ours} 0.363 bpp & 0.397 bpp \textbf{BPG} \\
\includegraphics[width=0.4\textwidth]{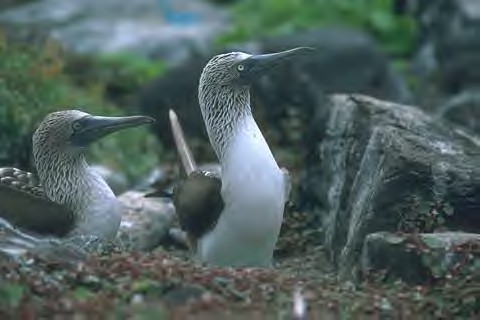}&
\includegraphics[width=0.4\textwidth]{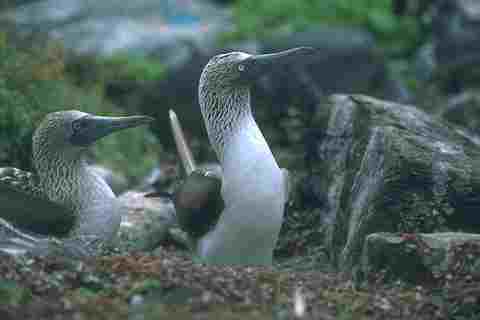}\\
\textbf{JPEG 2000} 0.369 bpp & 0.372 bpp \textbf{JPEG}

\end{tabular}
\vspace{-0.2cm}
\caption{\label{fig:vis_ex_B100_third}Our approach vs.\ BPG, JPEG and JPEG 2000 on the third and fourth image of the B100 dataset, along with bit rate.}
\end{figure*}

\end{document}